%% file: main.tex
\PassOptionsToPackage{dvipsnames}{xcolor}
\documentclass[]{kw_latex_template}

\usepackage{minitoc}
\usepackage{cleveref}
\usepackage{subcaption}
\usepackage{booktabs}
\usepackage{array}
\usepackage{multirow}
\usepackage{longtable}
\input{macro}

\title{GenericAgent: A Token-Efficient Self-Evolving LLM Agent via Contextual Information Density Maximization (V1.0)
}
\author{
Advantage AI Agent Lab (A3 Lab)
\thanks{This is a joint research lab with members from  Shenzhen Aquaintelling Technology and Fudan University} \\
}
\vspace{-0.5cm}
\contribution{See Author Contributions (Section ~\ref{sec:contributions}) for a full author list.}

\abstract{
\input{section/0abstract}
}
\begin{document}
\maketitle
\makeatletter\@thanks\makeatother

\section{Introduction}
\label{sec:intro}
\input{section/1intro}

\section{GenericAgent}
\label{sec:method}
\input{section/2method}

\section{Evaluation}
\label{sec:eval}
\input{section/4evaluation}

\section{Discussion}
\label{sec:discussion}

\input{section/6discussion}

\section{Related Work}
\label{sec:related}
\input{section/5related}

\section{Conclusion}
\label{sec:conclusion}
\input{section/7conclusion}

\clearpage

\bibliographystyle{unsrt}
\bibliography{main}
\clearpage
\section{Author Contributions}
\label{sec:contributions}
This project is led by \textbf{Jiaqing Liang} and \textbf{Yanghua Xiao}. The development of the GA and the preparation of this manuscript involve contributions from multiple authors across different aspects, including system design, experimentation, and writing. 
Below we detail the specific contributions of each author.

\noindent\textbf{Jiaqing Liang}: Assistant Professor of Fudan University and AI chief scientist of Shenzhen Aquaintelling Technology. \textbf{Led core system design, implemented the main codebase, and was responsible for project management}.

\noindent\textbf{Jinyi Han}: Led the manuscript design and writing process. Defined the paper structure, coordinated contributions across authors, and contributed to drafting, integrating, and refining the manuscript from the initial draft to the final version.

\noindent\textbf{Weijia Li}: Participated in the development of the GA system. Wrote Sections~\ref{sec:principle},~\ref{sec:autonomous-exploration}, and~\ref{sec:exp_browsing}, conducted the evaluation experiments therein, and designed Figures~\ref{fig:principle} and~\ref{fig:agent_loop}.

\noindent\textbf{Xinyi Wang}: Participated in the development of the GA system. Conducted evaluation experiments for Section ~\ref{sec:exp_memory} , contributed to the writing of Sections~\ref{sec:exp_memory} and \ref{sec:exp_browsing}.

\noindent\textbf{Zhoujia Zhang}: Participated in the development of the GA system. Conducted evaluation experiments for Section ~\ref{sec:self_evolution_capability}.

\noindent\textbf{Zishang Jiang}: Wrote Section ~\ref{sec:exp_task_completion}, ~\ref{sec:dimension4_tool_use_efficiency} and ~\ref{sec:self_evolution_capability}.

\noindent\textbf{Ying Liao}: Participated in the development of the GA system. Conducted evaluation experiments for Section~\ref{sec:dimension4_tool_use_efficiency} and generated figures for Section~\ref{sec:eval}.

\noindent\textbf{Tingyun Li}: Wrote Section ~\ref{sec:related} and the Appendix.

\noindent\textbf{Ying Huang}: Participated in the development of the GA system. Conducted evaluation experiments on RealFineBench for Section~\ref{sec:exp_task_completion}.

\noindent\textbf{Hao Shen}: Participated in the development of the GA system.

\noindent\textbf{Hanyu Wu}: Participated in the development of the GA system. Refined the writing of Section~\ref{sec:autonomous-exploration} and implemented the code for this module.

\noindent\textbf{Fang Guo}: Participated in the development of the GA system.

\noindent\textbf{Keyi Wang}: Conducted evaluation experiments on SOP-Bench for Section~\ref{sec:exp_task_completion}.

\noindent\textbf{Zhonghua Hong}: Conducted evaluation experiments on Lifelong AgentBench for Section~\ref{sec:exp_task_completion}.

\noindent\textbf{Zhiyu Lu}: Performed partial evaluation experiments for models in Section~\ref{sec:exp_task_completion}.

\noindent\textbf{Lipeng Ma}: Participated in the evaluation experiments in Section~\ref{sec:exp_task_completion} and contributed to the refinement of Section~\ref{sec:eval}.

\noindent\textbf{Sihang Jiang}: Contributed to the refinement of the overall manuscript and wrote Section~\ref{sec:discussion}.

\noindent\textbf{Yanghua Xiao}: Professor of Fudan University and chief scientist of Shenzhen Aquaintelling Technology. Provided overall project supervision and strategic guidance.
\appendix
\section{Appendix}
\input{section/appendix}
\end{document}

%% file: macro.tex
\usepackage{natbib}
\usepackage{latexsym}

\usepackage[utf8]{inputenc}    
\usepackage[T1]{fontenc}       
\usepackage{microtype}         

\usepackage{amssymb}           
\usepackage{amsmath}           
\usepackage{amsfonts}          
\usepackage{amsthm}            

\usepackage{booktabs}          
\usepackage{multirow}          
\usepackage{makecell}          
\usepackage{colortbl}          
\usepackage[dvipsnames]{xcolor}           

\usepackage{graphicx}          
\usepackage{tikz}              
\usepackage{pgfplots}          
\usepgfplotslibrary{groupplots} 
\usepackage[edges]{forest}     
\usepackage{subcaption} 

\usepackage{algorithm}         

\usepackage{wrapfig}           
\usepackage{caption}           
\usepackage{adjustbox}         

\usepackage{paralist}          

\usepackage{pifont}            

\usepackage{hyperref}          
\usepackage{url}               

\usepackage[toc,page,header]{appendix} 

\usepackage{xspace}            
\usepackage{nicefrac}          
\usepackage{fontawesome5}

\usepackage{hyperref}
\usepackage{url}
\usepackage{amssymb}
\usepackage{pifont} 
\usepackage{subcaption}
\usepackage{xspace}

\usepackage{enumitem}
\usepackage{algorithm}
\usepackage{xcolor}
\usepackage{algpseudocode}

\usepackage{amsmath, amsfonts}
\usepackage{nicefrac}
\PassOptionsToPackage{table}{xcolor}
\usepackage{booktabs}
\usepackage{colortbl}  
\usepackage{multirow}
\usepackage{adjustbox}
\usepackage{hyperref}
\usepackage{tabularx}
\usepackage{hhline}
\usepackage{makecell}
\usepackage{arydshln}
\usepackage{array}
\usepackage{colortbl}
\usepackage{wrapfig}
\usepackage{caption}
\usepackage{booktabs}       
\usepackage{amsfonts}       
\usepackage{nicefrac}       
\usepackage{setspace}

\usepackage{placeins}
\definecolor{light_origin}{RGB}{207, 117, 58}
\definecolor{cell_purple}{RGB}{76, 141, 187}
\definecolor{small_blue}{RGB}{27, 63, 125}
\definecolor{small_origin}{RGB}{94, 41, 16}
\definecolor{mypink}{HTML}{D92879}
\definecolor{linkblue}{RGB}{69, 97, 156}
\definecolor{linkorange}{RGB}{220, 120, 40}

\hypersetup{
    colorlinks,
    linkcolor={small_blue},
    citecolor={small_blue},
    urlcolor={small_blue}
}

\definecolor{light_purple}{RGB}{235,236,242}

\definecolor{light_blue}{RGB}{244,249,254}

\newcolumntype{B}{>{\columncolor{blue!4}}c}
\newcolumntype{d}{>{\columncolor{brown!4}}c}
\newcolumntype{q}{>{\columncolor{green!2}}c}
\newcolumntype{R}{>{\columncolor{red!4}}l}     
\newcolumntype{P}{>{\columncolor{purple!4}}c}   
\newcolumntype{Y}{>{\columncolor{yellow!4}}c}   

\definecolor{bgcolor}{RGB}{242, 243, 245} 
\makeatletter
\def\adl@drawiv#1#2#3{%
        \hskip.5\tabcolsep
        \xleaders#3{#2.5\@tempdimb #1{1}#2.5\@tempdimb}%
                #2\z@ plus1fil minus1fil\relax
        \hskip.5\tabcolsep}
\newcommand{\cdashlinelr}[1]{%
  \noalign{\vskip\aboverulesep
            \global\let\@dashdrawstore\adl@draw
            \global\let\adl@draw\adl@drawiv}
  \cdashline{#1}
  \noalign{\global\let\adl@draw\@dashdrawstore
            \vskip\belowrulesep}}

\tcbset{
  aibox/.style={
    width=\linewidth,
    top=2pt,
    bottom=2pt,
    colback=hjy_blue,
    colframe=black,
    center,
  }
}
\newtcolorbox{AIbox}[1][]{aibox,#1}
\definecolor{lightblue}{rgb}{0.22,0.45,0.70}%

\PassOptionsToPackage{table}{xcolor}
\usepackage{booktabs}
\usepackage{placeins}
\definecolor{light_origin}{RGB}{207, 117, 58}
\definecolor{cell_purple}{RGB}{76, 141, 187}
\definecolor{small_blue}{RGB}{27, 63, 125}
\definecolor{small_origin}{RGB}{94, 41, 16}
\definecolor{mypink}{HTML}{D92879}

\definecolor{hjy_blue}{HTML}{F2F6FD}
\tcbset{
    training_data/.style={
        colback=hjy_blue,      
        colframe=black,      
        sharp corners,       
        left=3pt, right=3pt, top=3pt, bottom=3pt,
        boxrule=1pt,
        fontupper=\normalsize,
        parbox=false,        
        enhanced,
        borderline={0.5pt}{0pt}{black}
    }
}

%% file: section/0abstract.tex
Long-horizon large language model (LLM) agents are fundamentally limited by context. As interactions become longer, tool descriptions, retrieved memories, and raw environmental feedback accumulate and push out the information needed for decision-making. At the same time, useful experience gained from tasks is often lost across episodes.
We argue that long-horizon performance is determined not by context length, but by how much \emph{decision-relevant} information is maintained within a finite context budget. We present \textbf{GenericAgent (GA)}, a general-purpose, \textbf{self-evolving} LLM agent system built around a single principle: \textbf{context information density maximization}. 
GA implements this through four closely connected components: a minimal atomic tool set that keeps the interface simple, a hierarchical on-demand memory that only shows a small high-level view by default, a self-evolution mechanism that turns verified past trajectories into reusable SOPs and executable code, and a context truncation and compression layer that maintains information density during long executions.
Across task completion, tool use efficiency, memory effectiveness, self-evolution, and web browsing, GA consistently outperforms leading agent systems while using significantly fewer tokens and interactions, and it continues to evolve over time.\\
\textcolor{black}{\faGithub}\ \textbf{Project:} \url{https://github.com/lsdefine/GenericAgent} \\

%% file: section/1intro.tex
The recent emergence of agentic systems such as Claude Code~\citep{claudecode}, OpenAI Codex~\citep{openai_codex}, and OpenClaw ~\citep{openclaw_rl} marks a qualitative shift in what is expected of a Large Language Model (LLM). 
Rather than serving solely as passive text generators, LLMs are increasingly deployed as goal-directed agents that operate through terminals, file systems, browsers, and external tools. 
This transition substantially expands their functional scope, but also introduces concrete systems-level challenges in context management~\citep{memgpt,context_as_tool} and experience accumulation~\citep{reflexion,expel,voyager,lifelongagentbench}. 
In particular, next-generation agents are required to continuously receive and solve tasks within a persistent environment, accumulating cross-task state and experience over time. 
Success therefore depends jointly on the reasoning capability of the underlying model and the system's ability to retain and reuse that experience effectively. 
Together, these requirements give rise to two fundamental challenges.

The first challenge is \textbf{context explosion.}
As the agent interacts with the environment, the prompt grows continuously as tool definitions, retrieved memories, intermediate observations, and raw environmental feedback accumulate across steps~\citep{memgpt,context_as_tool}.
This growth is not merely a token-consumption concern, it directly undermines reasoning quality. 
LLMs have finite effective attention~\citep{lost_in_middle,effective_context_length}. 
As irrelevant or outdated content occupies an increasing share of the context window, the model's ability to attend to decision-relevant information degrades.
Critical constraints are overlooked; intermediate states are confused with earlier ones; hallucinated facts emerge and compound through subsequent steps~\cite{lost_in_middle,multi_turn_lost,anthropic_context}. 
The core tension is that multi-step execution inherently requires accumulating context, yet the accumulated context itself becomes the primary source of failure. Without well-designed context management mechanisms, longer interaction does not yield better-informed decisions.

The second challenge is \textbf{effective experience accumulation and reuse}.
In long-horizon environments, critical knowledge, such as user preference, tool behaviors and effective action patterns, is not available at the outset. 
It emerges only through repeated trial and failure during actual task execution. 
This exploration is a natural and necessary process. 
The key question is whether the lessons learned can be retained and reused when similar tasks arise later. 
Without such a mechanism, agents repeat the same failure patterns across sessions~\citep{lifelongagentbench,mem0}. Successful strategies, once discovered, are forgotten upon context expiration.
Token expenditure scales linearly with task count, yet effective capability remains flat—a stagnation loop with no return on accumulated interaction. 

Existing agent frameworks largely fail to address this. 
Most treat each task episode as stateless, with no persistent memory across sessions~\citep{lifelongagentbench,rethinking_stateful_tool_use}. Even when retrieval-augmented memory is introduced, it typically stores raw logs rather than distilled, reusable operational knowledge~\citep{generative_agents,memorybank}. 
More critically, there is no feedback-driven refinement. Stale or incorrect memories are never updated, leading to silent degradation instead of improvement~\citep{rmm,telemem}.
These gaps prevent current agents from achieving self-evolution.

We propose \textbf{GenericAgent (GA)}, a self-evolving LLM agent system built around a fundamental principle: \textbf{maximizing contextual information density}.
The core view is that agent performance is not determined by context length alone, but by how much decision-relevant information is introduced, maintained, and updated within the limited context budget.

GA realizes this principle through four mechanisms that operate across the lifecycle of contextual information.
(1) A minimal atomic tool set reduces persistent tool overhead, preventing low-value interface information from occupying the context before task execution.
(2) A hierarchical memory mechanism selectively retains only verified and task-relevant knowledge, shaping what information persists over time.
(3) Self-evolution pipeline compresses interaction trajectories into reusable Standard Operating Procedures (SOPs), code, and skills, progressively transforming experience into compact and structured capability.
(4) A context truncation and compression mechanism actively manages historical information when the context exceeds its budget, ensuring the active context remains concise and task-relevant through layered truncation and compression.
Together, these mechanisms continuously reshape the context from a passive accumulation of information into a high-density, decision-centric representation.

We conduct comprehensive experiments on multiple benchmarks against representative agent systems, demonstrating that GA consistently achieves a superior efficiency, performance trade-off, reaching higher task completion at substantially lower token cost, with this advantage remaining stable across repeated runs.

The report is structured as follows. Section~\ref{sec:method} introduces the core design philosophy of GA and describes its key components, including the minimal atomic tool set, hierarchical memory architecture, self-evolution mechanism, and context truncation and compression. Building on these foundations, we present several emergent capabilities arising from their interaction, such as subagent dispatch, watchdog mechanisms, scheduled execution, and autonomous idle behavior. We then further examine higher-level capabilities enabled by GA’s minimalist architecture, including compositional capabilities and autonomous exploration. Section~\ref{sec:eval} provides a comparative evaluation against widely used agent frameworks on real-world tasks, focusing on both task success rate and token efficiency. The results highlight consistent advantages of GA in memory utilization, the self-evolution loop, and browser interaction design. Moreover, Section~\ref{sec:discussion} summarizes key findings for future agent design and reviews related work for each major component. We finally summarizes key findings for future agent design and reviews related work for each major component in Section~\ref{sec:related}.


%% file: section/2method.tex
\subsection{Design Principles}
\label{sec:principle}
\begin{AIbox}
\begin{center}\textbf{Context information density} is all a self-evolving LLM agent needs.\end{center}
\end{AIbox}

The performance of LLM-based agents is determined not merely by context length, but more fundamentally by context quality. Recent studies have identified three compounding failure modes that limit how effectively models can use their input. First, models exhibit pronounced positional bias when processing long sequences: relevant information placed in the middle of the context is significantly harder to retrieve than information near the beginning or end~\citep{liu2023lostmiddlelanguagemodels}. Second, irrelevant content does not simply remain unused; it can actively degrade performance by diverting attention away from decision-critical  evidence~\citep{shi2023largelanguagemodelseasily}. Third, the effective context length of LLMs falls far short of their nominal window size, meaning that part of the provided context is functionally inaccessible during generation~\citep{an2024doeseffectivecontextlength}.

These three phenomena reinforce one another in practice. As context grows longer, positional bias makes middle-positioned evidence harder to retrieve, irrelevant content competes for the model's limited effective attention, and the declining ratio of effective to nominal context length means that an increasing fraction of the prompt is simply wasted. The result is that, beyond a certain point, adding more context does not improve performance and may instead reduce it.

This has direct implications for agent system design. Existing agent frameworks that rely on monolithic prompt assembly often devote a disproportionate share of the effective context budget to scaffolding, control text, and marginally relevant interaction history. Rather than exposing only decision-relevant information, such strategies displace the evidence most needed for the current step, leaving the agent with more context in volume but less context in utility, while still increasing inference cost.

{From the perspective of context engineering, this problem is best understood in terms of two primary requirements and one secondary representational constraint:}
\begin{itemize}
    \item \textbf{Completeness.} {\small
    All information required for the current decision must be explicitly present in the context, preventing the model from relying on implicit assumptions or hallucinated inferences.
    }

    \item \textbf{Conciseness.} {\small
    Irrelevant and redundant information must be eliminated so that attention remains focused on decision-critical signals.
    }

    \item \textbf{Naturalness.} {\small
    The context should remain reasonably natural and semantically legible, especially when aggressive compression or artificial encodings make the representation harder for the model to use. This is an important but secondary constraint: in most agent settings, the dominant failure mode is not lack of natural phrasing itself, but the loss of completeness or conciseness.
    }
\end{itemize}

We argue that completeness and conciseness define the primary design space of context quality in LLM-based agents. Completeness ensures that the model has the information it needs, while conciseness ensures that this information is not diluted by distracting or low-value content. Naturalness still matters, but mainly as a constraint on how information is represented rather than as the main bottleneck in most agent failures. It helps avoid brittle encoding, overly telegraphic summaries, and artificial formats that the model may parse less reliably. Other commonly discussed dimensions can largely be understood within this framework. For example, recency contributes to both completeness and conciseness: the latest state is often essential for the current decision, while outdated information is typically no longer useful.  Relevance, rather than being an independent axis, is better understood as a cross-cutting concern: both completeness and conciseness presuppose a judgment of what matters, the former in deciding what must be included and the latter in deciding what should be removed.

\begin{figure*}[t]
    \centering
    \includegraphics[width=0.98\linewidth]{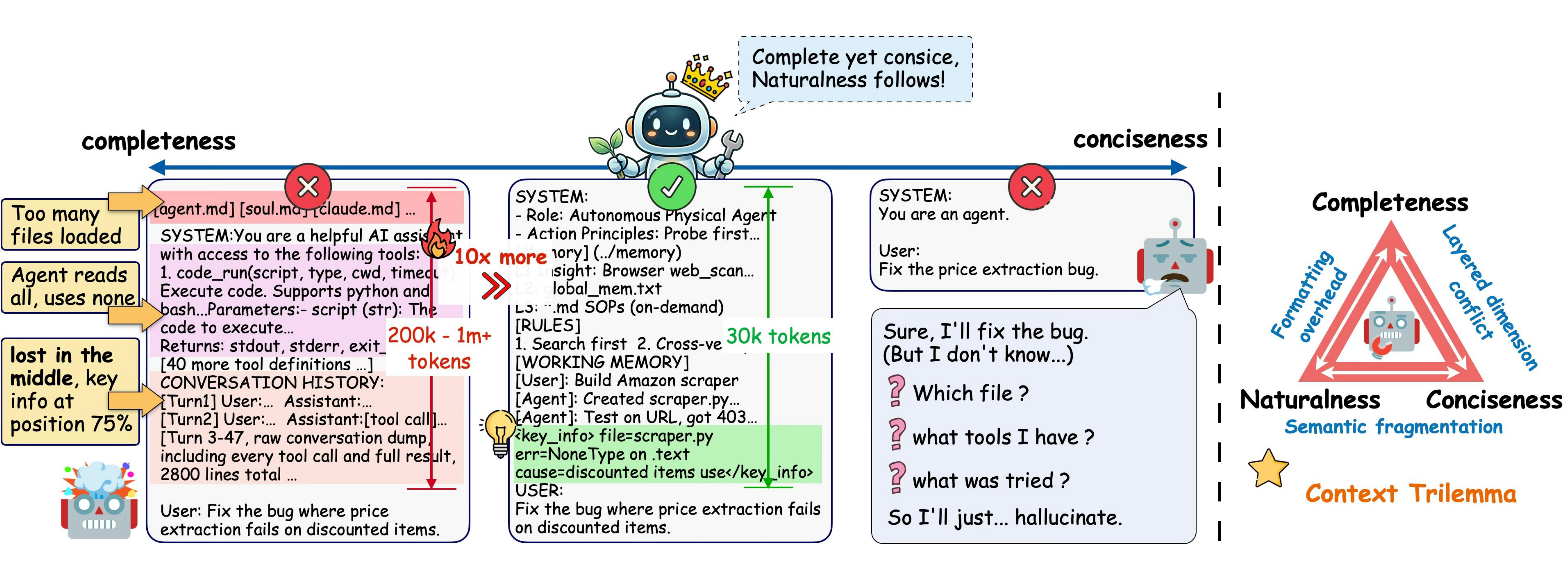}
    \caption{\textbf{Completeness and conciseness define the core trade-off in context design, while naturalness acts as a constraint on valid representations.} The left example is complete but overly verbose, obscuring key information. The right example is concise but incomplete, missing critical context. In GA, the context balances both, yielding a natural and effective representation.}
    \label{fig:principle}
\end{figure*}

\textbf{Critically, the main structural tension is between completeness and conciseness.} This is not merely a consequence of finite context budgets. Even if the context window were unbounded, a structural tension would remain. The tension is structural rather than merely budgetary for at least three reasons (as shown in Figure~\ref{fig:principle}):
\begin{itemize}
    \item Including more potentially relevant information improves completeness but weakens conciseness.
    \item Summarization and compression improve conciseness but risk omitting details needed for completeness.
    \item Naturalness can sharpen this trade-off in some cases, because highly compressed or artificial representations may be harder for the model to interpret, but this is typically a secondary effect rather than the dominant constraint.
\end{itemize}

The finite effective context window~\citep{an2024doeseffectivecontextlength} sharpens this tension into a hard practical constraint, but the underlying conflict is structural rather than merely resource-driven. In this sense, context engineering is best viewed not as an equal three-way trilemma, but as a constrained optimization problem centered on balancing completeness and conciseness, with naturalness acting as a supporting constraint on representation.

\textbf{This principle serves as the guiding idea behind the design of GA.} Rather than treating context as a passive byproduct of interaction history, GA systematically optimizes for information density at multiple layers of the system. Its memory hierarchy keeps a lightweight orientation layer visible while leaving richer facts and procedures to explicit reading, preserving conciseness without giving up access to deeper knowledge. Its browser interaction layer represents web pages through semantically structured observations rather than raw HTML, which improves readability while avoiding large amounts of low-value markup in context. Its self-evolution mechanism provides a way to consolidate successful experience into reusable memory and procedures, helping later tasks begin from more compact and task-relevant context. The following subsections describe how each mechanism instantiates this principle in detail.

\subsubsection{Tool Minimality}
\label{sec:tool_principle}
To maximize context information density before task execution begins, GA constrains its tool design to the minimal necessary set. Tool proliferation introduces system-level costs at two levels:
\begin{itemize}
    \item \textbf{At the prompt level}, each additional tool enlarges the tool schema and associated instructions injected into the context; in multi-turn interaction, this overhead accumulates and occupies effective context budget that should be reserved for task-relevant information.
    \item \textbf{At the policy level}, each additional tool enlarges the action space and increases ambiguity in tool selection. This makes planning more brittle, weakens the stability of tool-use patterns, and increases the likelihood of execution errors and unnecessary retries.
\end{itemize}
Tool minimality is therefore not merely an austerity choice; it is a better operating point that reduces both prompt overhead and decision complexity.

In practice, tool selection must satisfy two conditions: \textbf{atomicity}, which constrains each tool to an irreducible primitive capability, and \textbf{compositional generalization}, which allows complex behaviors to be realized through sequences of such primitives. GA obtains capability through composition rather than tool enumeration: a small set of atomic tools are designed as reusable primitives rather than task-specific interfaces, and complex behaviors emerge from their combination. Retained operational knowledge is therefore naturally expressed as reusable ways of sequencing and combining a small set of general primitives, rather than as new task-specific interfaces, making experience consolidation more lightweight. Tool minimality is thus not a restriction, but the core mechanism by which GA preserves general-purpose capability while reducing interaction overhead and creating favorable conditions for subsequent experience consolidation.

\subsubsection{Hierarchical Memory}
To continuously maintain context information density during task execution, GA adopts a systematic approach to memory organization. For a general-purpose agent, the main memory challenge lies less in storage itself than in controlling how much information remains in the active context. In conventional agent frameworks, prior interactions, intermediate states, and execution traces accumulate as the task unfolds, progressively consuming the context budget and burying decision-critical information beneath low-value content, thereby degrading the quality of reasoning at each step.

GA's core idea is to avoid treating all retained information as prompt-resident by default: only a small always-on layer remains visible, while deeper memory is accessed only when needed. This on-demand access principle naturally leads to a layered memory organization. The always-on layer is intentionally small, containing only lightweight orientation information, whereas richer factual knowledge, procedural knowledge, and historical interaction data are stored in deeper layers in archived or compressed form. These deeper memories enter the active context through on-demand retrieval rather than default injection. As a result, memory does not steadily displace the active-context budget required for the task at hand, preserving more of the context window for current reasoning and action.

The always-on layer can stay minimal because it needs to encode only the \emph{existence} of each knowledge category: the LLM itself serves as both compressor and decoder---when new knowledge is consolidated, it produces a near-minimal natural-language summary of the \emph{existence} under a word-limited requirement prompt, and at retrieval time it can follow any such compressed pointer to the relevant deeper layer through tool calls.

\subsubsection{Self-Evolution as Experience Consolidation}
To enable context information density to improve continuously across tasks over the long term, GA introduces a self-evolution mechanism. If hierarchical memory controls how past information is retained and accessed within a single task, self-evolution concerns whether past execution can remain useful beyond the episode in which it occurred. In real long-horizon environments, much valuable knowledge is not available at the outset, but emerges gradually through repeated interaction with a specific user's file system, workflows, and external services. Without mechanisms for consolidating such experience, agents cannot benefit from prior interactions, and must rediscover solutions from scratch on every similar task, incurring substantial and avoidable efficiency costs. \textbf{GA is designed to address this challenge by enabling experience consolidation across tasks, rather than treating each task as a fully isolated interaction.} 

Constrained by the context-quality trilemma discussed above, without any mechanism for retaining validated experience, the agent often faces a recurring dilemma on repeated tasks: either replay a longer exploratory process, which weakens conciseness, or act from a shorter but underspecified prompt, which weakens completeness. GA offers a way out of this trade-off. Rather than directly recording optimal action sequences, it provides a stronger starting point for future tasks, progressively improving reuse performance over repeated executions.  GA does not modify the base model directly; instead, it evolves by recording and updating the informational environment in which the model operates. Accordingly, memory management centers on selective consolidation, favoring information grounded in successful execution while filtering out volatile, weakly verified, or purely situational content.

This selective view also clarifies the relationship between self-evolution and the two mechanisms discussed above. Retained operational knowledge is naturally expressed as reusable ways of sequencing and combining a small set of general primitives rather than new task-specific interfaces, making experience consolidation more compact and lightweight, which is precisely where tool minimality creates favorable conditions for self-evolution. At the same time, consolidation does not reduce every past task to a single uniform reusable object; instead, drawing on the layered memory structure, the effective parts of past execution are preserved in forms that are smaller, more targeted, and potentially more reusable than the original trajectories. This is where hierarchical memory provides the organizational foundation for self-evolution.

\subsubsection{Context Truncation and Compression}
Many agent frameworks rely on extended context windows of up to 1M tokens, assuming that more context yields better reasoning. In practice, it is very expensive, but introduces more hallucinations. We refer to the effective ceiling as the \emph{hallucination-free context length}, which for current models is roughly an order of magnitude smaller. GA therefore targets a compact budget of under 30k tokens and invests in compression rather than expansion---packing higher-density information into a smaller window outperforms feeding diluted content into a larger one.

To actively maintain context information density during long-horizon execution, GA introduces a context truncation and compression mechanism. As the conversation progresses, accumulated context eventually exceeds the model's effective processing capacity; without active management, low-value historical content increasingly occupies the context budget needed for ongoing reasoning. Instead of permitting unbounded context growth, GA organizes historical information through layered management at multiple granularities. Tool outputs are first truncated with a head-tail policy to bound the size of individual messages. Redundant content in older messages is then compressed at the tag level to remove low-value fragments. When the global context budget is exceeded, the oldest messages are evicted in chronological order. Meanwhile, a continuously injected working-memory anchor prompt preserves task-critical information in the active context. Collectively, these mechanisms keep the active context compact and decision-relevant throughout long-horizon execution, preventing it from growing linearly with the number of interaction turns.

\begin{figure*}[t]
    \centering
    \includegraphics[width=0.98\linewidth]{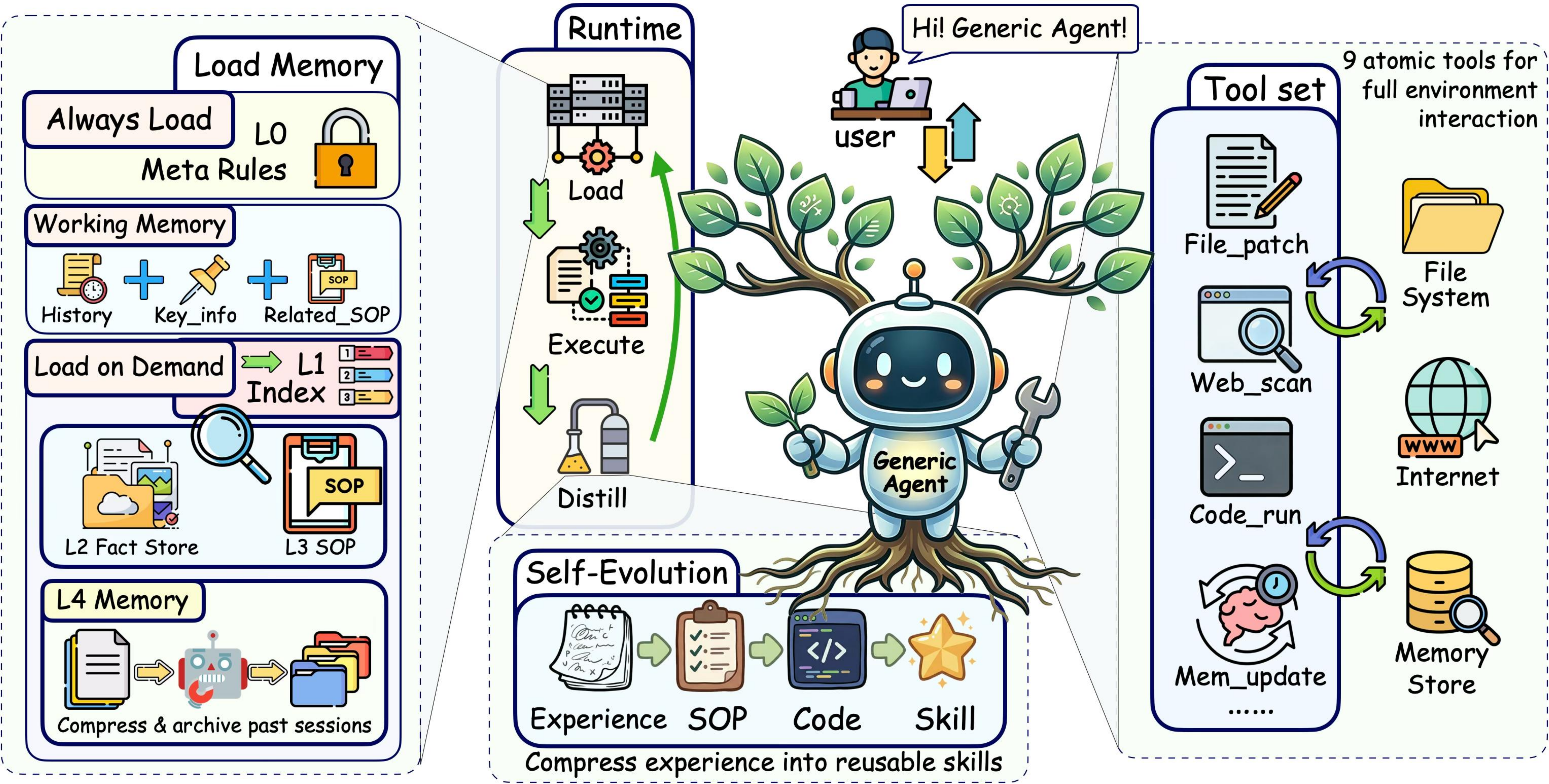}
    \caption{\textbf{Overall framework of GA.} GA follows a unified agent loop that constructs an execution context from the current task and relevant memory, generates outputs or tool calls, and updates the system through structured feedback.
It is built on four core mechanisms: a minimal tool set, hierarchical memory, reflection-driven self-evolution, and structured browser extraction.}
    \label{fig:agent_loop}
\end{figure*}

\subsection{Overview of GenericAgent}
\label{sec:overview}
Guided by the principles above, we develop GenericAgent, a general-purpose self-evolving LLM agent system designed to automate tasks within a local computing environment. It enables a compatible LLM backend to operate across browsers, terminals, file systems, keyboard and mouse inputs, screen-based perception, and mobile devices through configurable tool adapters. The system is model-agnostic: the underlying reasoning engine (e.g., Claude, GPT, Gemini) can be replaced without affecting execution logic, tool interfaces, or memory architecture. The overall architecture is illustrated in Figure \ref{fig:agent_loop}.

GA is organized around a unified agent loop that connects the model with the environment. At each step, the system combines global memory with the current task specification to form an execution context. The LLM processes this context and produces either direct outputs or tool calls. Tool execution is delegated to external modules, and the results are returned as structured signals that update the system state. This loop creates a continuous interaction between reasoning and environmental feedback, enabling tool-grounded iterative execution. After task completion, execution traces are compressed into structured long-term representations and stored in shared memory. This allows accumulated experience to improve future execution through compression and reuse rather than relying on unbounded computation.

On top of this loop, GA supports two execution modes. \textbf{Interact mode} handles user-initiated tasks such as code execution, information retrieval, and file operations. \textbf{Reflect mode} requires no user instruction; instead, it continuously monitors for environmental changes and automatically triggers the corresponding task when a specific condition or event is detected, analogous to the watchdog pattern described later. Both modes share the same agent loop and memory system, differing only in how they are triggered and how outputs are handled. To support long-horizon operation, GA defines explicit execution bounds that ensure controllability while preserving interruptibility and scope constraints.

\subsection{Core Components of GenericAgent}
\label{sec:core_design}
Building on the principles established in Section \ref{sec:principle}, we describe how GA realizes each of them through four concrete components: the atomic tool set, hierarchical memory management, self-evolution, and context truncation and compression.

\subsubsection{Minimal Atomic Toolset}
\label{sec:tools}

GA’s toolset adheres to a minimalist design: a small set of atomic primitives that can be composed to solve a wide range of tasks. This section describes which tools are provided, how they are defined and invoked, how their boundaries are enforced, and how they are optimized for efficient decision-making.

\textbf{Tool composition and capability coverage} GA exposes nine atomic tools, organized into five capability classes.
The \textbf{File Operations} class includes \texttt{file\_read}, \texttt{file\_patch}, and \texttt{file\_write} for eading, precise editing, and block writing.
The \textbf{Code Execution} class contains \texttt{code\_run}, which executes Python or Bash within a controlled runtime.
The \textbf{Web Interaction} class includes \texttt{web\_scan} and \texttt{web\_execute\_js} for low-cost page inspection and precise browser actions.
The \textbf{Memory Management} class includes \texttt{update\_working\_checkpoint} and \texttt{start\_long\_term\_update} for short-term context maintenance and long-term memory distillation.
The \textbf{Human-in-the-loop} class is \texttt{ask\_user}, used when user decisions are required.
Each tool maintains a single responsibility, and these tools form a complete loop covering state observation, action execution, context preservation, and intervention requests.
This tool set directly instantiates the atomicity and compositional generalization principles discussed in Section~\ref{sec:tool_principle}. Each tool is restricted to a single, indivisible capability with no functional overlap. Consequently, complex tasks are executed by combining these basic primitives, rather than by introducing new, specialized interfaces.

\textbf{Declaration, invocation, and dispatch.} GA represents each tool as a verifiable schema contract and routes execution through a unified dispatcher.
Each tool defines a name, a description, and a set of parameters described by JSON Schema, all under a consistent type function format.
At runtime, this schema is injected into the interface, enabling GA to produce structured \texttt{tool\_use} calls with explicit names and arguments.
The dispatcher maps each call to a local executor and manages pre- and post-execution handling together with the returned results, ensuring a clear separation between declaration, invocation, and execution.

\textbf{Capability boundaries and execution control.}  GA enforces a clear permission hierarchy via the injected toolset. Each tool’s capability scope is strictly bounded and non-overlapping, avoiding the ambiguity and redundancy associated with overlapping functionality. With read-only tools, GA can inspect and reason about state but cannot modify it; with patch tools, GA can edit text, subject to strict matching and parameter constraints; with execution and web-control tools, GA can trigger real system actions. All interactions return structured outputs through the dispatcher, which improves controllability, reduces operational risk, and ensures full traceability of every action.

\textbf{Tool specific optimization for execution efficiency.} GA further optimizes each atomic tool for execution efficiency and token economy.
\texttt{file\_read} supports segmented reads (\texttt{start}/\texttt{count}), keyword anchoring, and line-number output, so GA can inspect only the relevant regions.
\texttt{file\_patch} enforces unique \texttt{old\_content} matching and fails fast on zero or multiple matches, reducing the risk of silent multi-location edits.
\texttt{web\_scan} incorporates a layout-analysis algorithm that clones the live Document Object Model (DOM), computes per-element visibility, classifies regions as main or non-essential through overlay and partition analysis, and removes covered or hidden elements before serialization, reducing token cost by an order of magnitude compared to raw DOM output.
\texttt{web\_execute\_js} returns action results plus page-change observations, so many workflows proceed without another full scan.
\texttt{code\_run} is restricted to one invocation per turn, ensuring each result is observed before the next action is planned.
The goal of these optimizations extends beyond merely compressing individual tool outputs. By providing more precise and complete feedback at each step, they reduce redundant probing and retries in subsequent turns. Ultimately, this lowers overall token consumption across multi-turn interactions and maximizes context information density.


Theoretically, the agent could accomplish any task using only the \textbf{code\_run} tool. By executing custom scripts, it has the ability to replicate the exact functions of every other tool.
Therefore, the remaining eight tools are not designed to expand the agent's capabilities. Instead, they serve as specialized shortcuts to reduce the agent's decision-making cost and operational overhead. Rather than forcing the agent to write code from scratch for every minor action, this carefully designed toolset acts as a \textbf{harness}, keeping the agent focused on efficient problem-solving.

\subsubsection{Hierarchical Memory Architecture}
\label{hierarchical_memory}
To maximize context information density during task execution, GA organizes memory into three functionally distinct types rather than a single undifferentiated store.
\textbf{Working memory} is continuously injected across turns, carrying only the minimal necessary task state: current objectives, constraints, and progress. This allows the agent to maintain execution continuity without reconstructing the full context each time.
\textbf{Always-on memory} remains persistently visible but is deliberately compressed to the lightest possible orientation information. It provides only the basic memory layout and navigation index, rather than loading complete knowledge into the context by default.
\textbf{Long-term memory} is kept outside the default context and managed via an explicit post-task consolidation process; only verified and stable knowledge is written back, ensuring that historical accumulation does not automatically translate into continuous context growth.
This division directly embodies the principle of context information density maximization: each type of memory appears in the active context only at the necessary moment and at the necessary level of granularity.

To realize this design at the implementation level, GA defines a four-layer memory architecture.
\textbf{(1) L1: index layer.}
It stores compact pointers, including high-frequency entry points, keyword mappings, and a small set of hard constraints, serving as the core content of always-on memory for fast navigation and efficient memory routing.
\textbf{(2) L2: fact layer.}
It stores verified and stable factual information that remains valid over long periods. Admission is strictly controlled: information is written into L2 only after being validated through execution and proven reusable across tasks, deliberately excluding transient states, one-time events, and unverified hypotheses.
\textbf{(3) L3: SOP layer.}
It stores reusable procedural knowledge, including task workflows, preconditions, key execution steps, common failure cases, and corresponding debugging or recovery strategies.
\textbf{(4) L4 : raw session archive layer.}
Unlike L1--L3, L4 is not designed for frequent direct injection into the runtime context; its role is persistence and traceability, storing historical execution sessions so the agent can reconstruct prior trajectories, audit past decisions, and recover task context when needed.
Within this hierarchy, L1 constitutes the always-on memory, L2 and L3 together make up long-term memory, and L4 provides durable storage for persistence and traceability.

To manage the four-layer architecture, GA introduces a global meta-memory layer.
The meta-memory layer defines the overall memory map, core rules, and update boundaries, providing the model with a shared frame of reference before execution so that it understands how memory is organized, what each layer is for, and how updates should be handled, thereby reducing arbitrary writes, historical misreads, and cross-task leakage.
The full meta-SOP content is loaded on demand via file reads rather than always preloaded, keeping the always-on layer lightweight.
In practice, GA injects only the meta-memory and L1 index layer by default, following an L1$\rightarrow$L2/L3 routing chain to retrieve deeper factual or procedural knowledge only when needed, excluding irrelevant content from the active context. This routing is enforced through tool calls and prompt policy rather than a hardcoded mandatory pipeline.

For long-term consolidation, GA uses a triggered commit mechanism instead of immediate writing.
When information is identified as potentially valuable, it enters a validation stage where its usefulness and reusability are checked before being committed.
Only verified information is written into L2 or L3 through small, incremental updates, with the L1 index updated accordingly, preventing memory from growing in an uncontrolled manner over time.

A critical design invariant is that L1 remains bounded even as L2 and L3 expand. Each L1 entry records only the \emph{existence} of a knowledge category rather than its substantive content. Consequently, new entries are introduced only when genuinely new categories arise, and the overall description length of L1 approaches the Kolmogorov complexity of the categorical structure of the knowledge set. This extreme degree of compression is feasible because the LLM itself serves as the decoder: once it infers that a relevant capability or fact exists, it can expend tool calls and tokens to retrieve the full content from deeper layers, so a minimal existence-level signal is sufficient for accurate routing.

\subsubsection{Self-Evolution Capability in GA}
\label{sec:self-evolution}
GA implements self-evolution as an explicit and transparent process rather than an accidental phenomenon of reasoning. Achieving this requires clearly defining what evolves, how knowledge accumulates, how quality is controlled, and how the evolutionary trajectory is maintained.

\textbf{What evolves: strategy, not tools.}
GA separates a fixed tool layer from an evolving knowledge layer.
While the tool interface and user interactions work for any task and stay the same during operation, all task-specific capabilities are stored in SOP files and reusable scripts. The agent can easily read, create, and modify these using its own tools.

This separation ensures that learning new tasks doesn't interfere with existing skills, allowing knowledge to grow safely over time. Over multiple sessions, real-world feedback helps refine the SOPs. Common subtasks naturally evolve into stable, reusable scripts, upgrading the agent's knowledge from plain-text instructions to executable code.

\textbf{How knowledge accumulates: hierarchical memory.}
Building on the hierarchical memory described in Section~\ref{hierarchical_memory}, GA ensures that knowledge gained during one session is immediately available in future ones. The active L1 index automatically tracks new L2 facts and L3 procedures. This eliminates the need for manual updates. Consequently, every successfully completed task permanently expands the agent's practical capabilities.

\textbf{How quality is controlled: selective consolidation.}
GA saves raw action traces at the lowest memory level (L4), but it does not automatically promote these traces to higher, long-term memory levels, such as L2 or L3. Instead, reusable L3 procedures are generated only through explicit consolidation steps. These steps are triggered at meaningful milestones, such as successfully completing a subgoal or recovering from a system error.

During this consolidation phase, the agent only retains information that has been strictly verified through successful tool execution.
Guesses, temporary intermediate states, and failed decision branches are systematically discarded under the ``No Execution, No Memory'' rule. This strict filtering prevents memory pollution, ensuring that the accumulated knowledge remains reliable and practically reusable in the future.

\textbf{How the evolutionary trajectory is maintained: failure escalation.}
To prevent the agent from getting trapped in endless loops of incorrect actions, which would degrade performance rather than improve it, GA introduces a staged escalation mechanism for handling failures.

When a task fails, GA handles the error through a clear, three-step recovery process:
First, it analyzes the immediate error message to make a small, localized adjustment and tries again.
Second, if the failure persists, the agent abandons its current approach. It is forced to either switch to a completely new strategy or search the environment for missing information.
Finally, if all automated attempts fail, the agent pauses and requests human intervention.
This structured design prevents the agent from blindly repeating mistakes. By triggering progressively stronger corrective actions, it ensures the agent breaks out of dead ends and keeps its long-term learning on the correct path.

\subsubsection{Context Truncation and Compression}
\label{sec:context-truncation}
The underlying model that GA relies on operates within a finite context window, and the cumulative conversation length must be kept below this limit at all times. Since GA cannot directly obtain precise token counts, context budget management uses a character-domain heuristic. Specifically, $C_{\mathcal{H}}$ denotes the \textbf{total character length} of all messages in the current conversation history, and $B$ denotes the \textbf{character budget upper bound} converted from the token budget; the corresponding compression or eviction mechanism is triggered when $C_{\mathcal{H}} > B$:

\begin{equation}
  C_{\mathcal{H}} = \sum_{m \in \mathcal{H}} \mathrm{len}(m),
  \qquad
  B = \alpha \cdot W_{\text{tokens}},
  \quad \alpha \approx 3 \;\text{chars/token},
  \label{eq:char-budget}
\end{equation}

\noindent
where $\mathrm{len}(m)$ denotes the character length of message~$m$
after JSON serialisation, $W_{\text{tokens}}$ is the configured token
window, and $\alpha$ is the empirical ratio used to approximate the
token budget as a character budget. This heuristic has a certain margin
of error. For ASCII-dominated content, each token corresponds to roughly
4~actual characters, so the $\alpha{=}3$ setting slightly underestimates
character efficiency, causing mildly early eviction---a conservative but
safe failure mode. For CJK content the situation reverses: each character
typically consumes 1--2~tokens, so the $\alpha{=}3$ ratio substantially
underestimates actual token usage, risking delayed eviction and potential
context overflow.

GA uses four distinct context-trimming mechanisms to manage context length, ranging from fine-grained edits to broad removals. First, \textbf{tool-output truncation} controls the size of individual messages. Second, \textbf{tag-level compression} removes low-value fragments from older entries. Third, \textbf{message eviction} deletes the oldest content entirely when the overall token budget is exceeded. Finally, a \textbf{working-memory anchor} ensures that critical task information remains visible even after prior messages are evicted.
Together, these mechanisms prevent the active context from growing linearly as the interaction continues. They ensure that the limited context budget is always strictly focused on information directly relevant to the current task.

\paragraph{Stage 1: Tool-output truncation.}

Each tool handler limits its return value \emph{before} adding it to the message history. 
If the output exceeds a predefined character threshold ($L$), the system retains only the first and last $L/2$ characters, replacing the middle section with an ellipsis.
Table~\ref{tab:tool-truncation}
lists the thresholds for outputs that enter the LLM context.

\begin{table}[h]
\centering
\caption{Per-tool truncation thresholds for content entering the LLM context.}
\label{tab:tool-truncation}
\begin{tabular}{lr}
\toprule
Tool / mode & $L$ (chars) \\
\midrule
\texttt{code\_run}                          & 10\,000 \\
\texttt{web\_execute\_js}$^{\dagger}$       &  8\,000 \\
\texttt{web\_scan} (\texttt{text\_only})    & 10\,000 \\
\texttt{web\_scan} (HTML)$^{\ddagger}$      & 35\,000 \\
\texttt{file\_read}                         & ${\sim}$1\,280/line; 20\,000 total \\
\bottomrule
\multicolumn{2}{l}{\footnotesize $^{\dagger}$\,With \texttt{save\_to\_file},
  the full output is written to disk; only a short preview enters the history.} \\
\multicolumn{2}{l}{\footnotesize $^{\ddagger}$\,DOM-level character budget
  (subtree pruning), not head--tail truncation.}
\end{tabular}
\end{table}

\paragraph{Stage 2: Tag-level compression.}
Older messages tend to accumulate redundant text inside XML tags (such as reasoning traces, tool invocations, and working-memory snapshots). 
To reduce computational overhead, the system runs a compression pass roughly every five turns. This pass handles redundancy in two ways: 
(i) \textbf{Placeholder replacement}: Repeated working-memory blocks (like (\texttt{<history>}, \texttt{<key\_info>})) are replaced with a short placeholder, as only the most recent snapshot is actually needed.
(ii) \textbf{Windowe truncation}: Content inside reasoning and tool tags (like (\texttt{<thinking>},
\texttt{<tool\_use>}, \texttt{<tool\_result>}) is truncated, keeping only an $\sim$800-character window (the beginning and the end of the text).
To ensure the model always has the latest context, the 10 most recent messages are exempt from this compression. The five-turn interval also helps with efficiency: the unchanged older messages result in prompt-cache hits in roughly 80\% of turns.

\paragraph{Stage 3: Message eviction.}
When a new message causes the total history length ($C_{\mathcal{H}}$) to exceed the character budget ($B$), the system triggers an eviction process. First, it reruns the Stage 2 compression using stricter rules, exempting only the 4 most recent messages. Next, it removes the oldest messages (FIFO order) until the history size drops below 60\% of the total budget, leaving sufficient room for future turns.

After eviction, the newly exposed oldest message is cleaned up to maintain API consistency (for example, converting orphaned tool-result references into plain text). Importantly, evicted content is not completely lost: the Stage 4 anchor preserves essential task states, and the Stage 2 compression ensures that even a $<$30k context window can hold dozens of historical turns.

\paragraph{Stage 4: Working-memory anchor prompt.}
After every tool invocation, an anchor prompt is automatically attached to the next user message. This anchor contains: (i)~the 20 most recent one-line turn
summaries (each compressed to roughly one hundred characters),
(ii)~the current turn number, and (iii)~a persistent \texttt{key\_info}
block maintained by the agent via
\texttt{update\_working\_checkpoint}.  

Because this anchor is injected into every new user message, Stage 2 compression automatically replaces the older copies with placeholders. This ensures only the newest version carries the full data payload. Once older messages are evicted in Stage 3, this anchor becomes the sole source of long-term memory, keeping task-critical information active in the context.

\paragraph{Auxiliary: tool-schema elision.}
When using the text-protocol path, if a tool's definition hasn't changed from the previous turn, its full schema is removed from the prompt and replaced with a brief natural-language reminder. To prevent the model from forgetting the exact tool formats over time, the full schema is periodically re-sent—either after a fixed number of turns or when the active prompt gets too long. (Note: This optimization is not used in the native API path, which requires sending complete tool definitions on every single call.)

\section{From Minimal Architecture to Emergent Capabilities}
\label{sec:Emergent_capability}
\subsection{Minimal Architecture}

GenericAgent exhibits architectural minimality in two dimensions: \textbf{code minimality} and \textbf{interface minimality}.

At the code level, GA is exceptionally lightweight. The core codebase contains only about 3,300 lines of code, with the central Agent Loop running on a mere 92 lines.
This stands in sharp contrast to comparable systems-OpenClaw's codebase is approximately 530,000 lines, more than 160 times larger than GA's. 
Rather than relying on heavy module dependencies or complex tool-registration infrastructures, 
GA's core logic is fully expressible within a minimal codebase, making the system easier to maintain, debug, and extend.

At the interface level, GA exposes itself as a self-hosted CLI program, making the command 
line not a wrapper around an internal platform, but the native 
execution surface of the system. Tasks can be launched with arguments or kept running in the background. GA therefore does not require plugin frameworks, event buses, or dedicated orchestration layers to support higher-level behavior. All interactions with GA—whether task submission, progress monitoring, or runtime intervention—are handled through this single unified CLI interface.

This extreme minimality is not a functional compromise, but a deliberate design choice that unlocks greater power. Because the architecture is stripped down and the interface is purely CLI-based, the system does not need new architectural layers to support advanced behaviors. Instead, features like \textbf{Subagent Dispatch} and \textbf{Reflect Mode} emerge naturally. For instance, because GA operates as a standalone CLI tool, an agent can spawn a subagent simply by executing a standard terminal command. Complex AI behaviors are thus built from a single, elegant system primitive rather than layered engineering complexity.

\subsection{Compositional Capabilities}
\label{sec:compositional_capability}
A recurring pattern in agent framework design is to introduce dedicated subsystems for each new capability—for example, a subagent manager for multi-agent coordination, or a dedicated listener daemon for event handling. 
GA takes a fundamentally different approach: once the agent is exposed as a standard CLI program, two important higher-level capabilities emerge naturally from this single primitive, without any extension to the core architecture.

\textbf{Subagent Dispatch.}
Once the agent can be invoked programmatically via CLI, subagents follow naturally. 
To handle complex subtasks, a parent agent simply runs a standard terminal command to launch multiple GA instances in the background. In this framework, a ``subagent'' is not a special object, nor does it require a dedicated manager. Parent and child are simply standard, independent processes communicating through the same interaction protocol.

Crucially, context isolation is naturally guaranteed: each instance runs in its own memory space and maintains its own conversation history, ensuring that subtasks never interfere with one another. For highly parallel tasks, this creates an elegant map-reduce workflow: the parent prepares independent inputs, spawns one subagent CLI process per input, and seamlessly merges the results once all child processes finish.

\textbf{Reflect Mode.}
Similarly, once the CLI can be invoked programmatically, daemon-like behavior emerges naturally. 
GA's Reflect Mode requires no user 
instruction; instead, it continuously monitors for environmental changes and automatically triggers the corresponding task once a specific condition is detected. 
The mechanism is straightforward: a lightweight script periodically checks conditions. When a rule is triggered, the script simply dispatches the returned string as a standard task to the GA CLI. This strictly separates the triggering logic from the execution logic: the external script decides when to create work, while the stable core runtime focuses entirely on how to execute it.

Watchdog and Scheduled Task are two concrete applications of this Reflect Mode:
\begin{itemize}
    \item \textbf{Watchdog:} Monitors for changes in the environment (e.g., a new file or an error log) and triggers GA immediately upon detection.
    \item \textbf{Scheduled Task:} Uses time-based rules to generate GA tasks at specific intervals or exact times.
\end{itemize}
Both share the exact same underlying mechanism and only differ in how the external trigger script is written. Because these rules are maintained as standalone scripts outside the agent, developers can update them at runtime without ever needing to restart the core agent process.

The common thread across these two capabilities is that neither required extending the core loop. Both emerge from the CLI as a single primitive, and the core runtime remains stable throughout. This reflects the deeper value of architectural minimality: a sufficiently simple system can support the widest range of behavioral compositions at the lowest possible cost.

\subsection{Autonomous Exploration Capability}
\label{sec:autonomous-exploration}

In real long-horizon environments, relying solely on user interaction
to accumulate skills is fundamentally inefficient.
User instructions are usually sparse and focused on immediate tasks, meaning they cannot systematically cover the full range of what the agent is capable of doing. Therefore, autonomous learning is a necessary feature for the continuous evolution of the agent.

The two capabilities established in Section~\ref{sec:compositional_capability}
together make this possible without any new architectural machinery.
Subagent dispatch provides the execution substrate: the agent can
programmatically launch child instances to handle exploration tasks in
parallel. Reflect Mode provides the trigger substrate: it continuously
monitors for idle conditions and fires an exploration task the moment
one is detected, with no user instruction required. Autonomous
exploration is therefore not a new subsystem, but the natural result of
combining these two primitives---the dispatcher becomes the agent
itself rather than the user. What remains to be defined is the internal
logic that decides \emph{what} to explore and \emph{how} to evaluate
the result; the rest of the execution path is already in place.

GA organises accumulated skills into a persistent \emph{skill tree}.
The skill tree is a two-level map: \emph{categories}
(e.g.\ \texttt{web\_automation}, \texttt{data\_processing}) contain
named \emph{skills}, each recording its associated tool scripts and a
monotonically increasing usage counter. 
This skill tree serves a dual purpose: it acts as a global index of the agent's current capabilities, and it is the core dataset driving its autonomous exploration decisions.

\textbf{Trigger Mechanism.}
Autonomous exploration can be initiated in two ways. In \emph{task
mode}, an external orchestrator (e.g.\ a cron job) submits a one-shot
task via a file-system mailbox and polls for results. In \emph{reflect
mode}, a user-supplied callback is polled at a fixed interval; when it
returns a non-empty string, that string is injected as a new prompt.
The default trigger fires every six minutes with a fixed prompt
directing the agent to consult its exploration procedure. Both modes are concrete instantiations of the Reflect Mode primitive
introduced in Section~\ref{sec:compositional_capability}, and require
no extension to the core architecture.

\textbf{Exploration Task Generation.}
A core question in autonomous exploration is determining what to
explore. When no pending task list exists, the agent enters planning
mode. The curriculum planner inspects the current skill tree and
scores task candidates along four dimensions using a weighted sum:
\begin{equation}
  S(t) \;=\; w_b\,B(t) \;+\; w_d\,D(t) \;+\; w_u\,U(t) \;+\; w_i\,I(t)
  \label{eq:task-score}
\end{equation}
\noindent where $B$ (\emph{breadth}) rewards filling under-populated
skill categories, $D$ (\emph{depth}) rewards enhancing frequently-used
skills, $U$ (\emph{utility}) captures estimated practical value, and
$I$ (\emph{innovation}) favours novel techniques or domains. Each
dimension is scored on a 1--10 scale.

Breadth and depth are computed directly from skill-tree statistics:
\begin{equation}
  B(t) = 10 \times \max\left(0,\, 1 - \frac{|S_c|}{\bar{S} + 1}\right),
  \label{eq:breadth}
\end{equation}
\begin{equation}
  D(t) = 10 \times \frac{u(t)}{u_{\max} + 1},
  \label{eq:depth}
\end{equation}
\noindent where $|S_c|$ is the number of skills in the target
category, $\bar{S}$ is the mean skill count across all categories,
$u(t)$ is the usage count of the target skill, and $u_{\max}$ is the
maximum usage count across all skills. The breadth formula rewards
filling categories that are below average, while the depth formula
prioritises enhancing frequently-invoked capabilities. Utility and
innovation are estimated by the LLM on the same 1--10 scale.

The weights are initially set to $(w_b, w_d, w_u, w_i) = (0.3, 0.2,
0.3, 0.2)$, prioritising breadth and utility. The resulting task list
must span at least four distinct skill categories, preventing narrow
concentration. Planning and execution are strictly separated: the
planner produces a task list and immediately yields control; execution
resumes in the next invocation.

\textbf{Execution and Consolidation.}
Each autonomous task proceeds through a fixed sequence:
\emph{context loading} (retrieve recent history and pending tasks),
\emph{skill search} (query the skill tree for relevant tools and
identify gaps),
\emph{execution} (research, prototype, and validate within a
sandboxed temporary directory),
\emph{report writing} (a Markdown document with machine-readable
metadata tags for automatic skill-tree updates),
\emph{skill consolidation} (an atomic operation that archives the
report, updates the skill tree, and increments usage counters), and
\emph{task-list advancement}.
Importantly, even failed experiments are thoroughly documented. This ensures that future planning rounds can read the history and avoid repeating the same dead ends. As a strict security measure, all generated files are confined to the temporary directory. Access to system secrets or core source code is unconditionally prohibited.

Furthermore, the system maintains a separate lightweight log that records three specific types of information: observed errors paired with their corrections, explicit user preferences, and verified success patterns. These entries are automatically injected into the system prompt to guide the agent's future behavior and improve its decision-making.

\textbf{Exploration Quality Assessment.}
The second core question is evaluating exploration quality. After each
batch of autonomous tasks, a \emph{reflection-based adaptation}
mechanism adjusts the scoring weights based on actual usage outcomes.
For each completed task~$t$, if its predicted score $S(t) > 8.0$ but
actual usage $u(t) < 3$ within 30~days, the weight of the dominant
dimension in $t$'s score vector is reduced by 10\%. Conversely, if
$S(t) < 5.0$ but $u(t) > 5$, the corresponding weight is increased
by 10\%. After these adjustments, all weights are re-normalized to ensure they sum to 1. 
This feedback loop allows the system to automatically discover which dimensions best predict practical value in the user's specific workflow. As a result, GA can continuously correct its future exploration decisions without requiring any manual tuning.


\textbf{Limitations.}
Several limitations merit discussion:
\begin{itemize}
    \item \textbf{Context Length Constraints:} The 30-round execution cap means highly complex research tasks may span multiple sessions. Currently, continuity between these sessions is maintained only through written reports and task-list annotations.
    \item \textbf{Unverified Adaptation:} The reflection-based weight adjustment is a preliminary design. It has not yet accumulated enough long-term data to definitively prove its effectiveness across diverse, real-world user workflows.
    \item \textbf{Manual Log Curation:} The self-improvement log (which records errors and preferences) currently relies on manual user curation, limiting how much it can autonomously influence the agent's core behavior.
    \item \textbf{Manual Tree Maintenance:} Advanced skill tree management, such as merging redundant categories, deprecating outdated tools, or restructuring the tree topology, remains entirely manual.
\end{itemize}

%% file: section/4evaluation.tex
\input{tables/main_result}
\input{tables/tool_inventory_overview}


In this section, we evaluate the GenericAgent to systematically analyze its system behavior, resource management, and core mechanisms. Our evaluation is structured around five complementary dimensions:
\begin{enumerate}
    \item \textbf{Task Completion and Token Efficiency:} We measure overall success rates and token consumption across diverse benchmarks to quantify the GA's fundamental execution capability and operational cost.
    \item \textbf{Tool-Use Efficiency:} We analyze the minimal atomic-tool design to assess how a restricted tool space impacts the ability to resolve complex workflows and the resulting interaction overhead.
    \item \textbf{Memory System Effectiveness:} We investigate the memory management mechanisms to examine their function in long-term fact retention, performance dynamics across repeated tasks, and context size control.
    \item \textbf{Self-Evolution Capability:} We evaluate the self-evolution pipeline to observe the process and effects of compressing historical interaction trajectories into reusable Standard Operating Procedures (SOPs) and executable code.
    \item \textbf{Web Browsing Capability:} We use open-ended web tasks to test the framework's end-to-end navigation, multi-hop retrieval, and multi-step execution in a dynamic, unstructured real-world environment.
\end{enumerate}

\subsection{Task completion and token efficiency}
\label{sec:exp_task_completion}
This section evaluates the fundamental execution capability and resource consumption of the agent systems. The primary objective is to investigate the baseline performance-cost trade-off---specifically, how efficiently a system translates token consumption into successful task execution. By measuring accuracy alongside context usage across diverse scenarios, this experiment aims to quantify the operational cost of maintaining high performance and verify whether the system's underlying context management effectively reduces token overhead without compromising task reliability.

\subsubsection{Setup}

\paragraph{Baseline.}
To contextualize the performance of \textbf{GA}, we establish a comparative evaluation against three representative agent systems: (1) \textbf{Claude Code} (an industry-leading proprietary agent), \textbf{OpenClaw} (a robust open-source multi-tool framework), and \textbf{CodeX} (a widely adopted execution baseline). Furthermore, to prevent model-specific biases and ensure the generalizability of our findings, the evaluations are conducted across a diverse spectrum of cutting-edge LLMs. The backbone LLMs deployed include \textit{Claude 4.6 Sonnet} and \textit{Claude 4.6 Opus} for top-tier proprietary capabilities, alongside \textit{MiniMax M2.7} and \textit{GPT-5.4}, covering different architectural paradigms and reasoning capacities.

\paragraph{Benchmark.}
The evaluation is conducted on three distinct benchmarks to assess distinct facets of agent capabilities under complex, real-world constraints:
(1) \textbf{SOP-Bench}~\citep{sopbench}: This benchmark evaluates the execution of multi-step SOPs, measuring the agent's instruction-following and procedural reasoning capabilities.
(2) \textbf{Lifelong AgentBench}~\citep{lifelongagentbench}: This benchmark consists of sequential tasks with explicit cross-task dependencies, designed to evaluate the agent's ability to maintain state and manage context over continuous interactions.
(3) \textbf{RealFin-Benchmark}~\citep{realfin}: This benchmark comprises financial workflows, used to assess domain-specific understanding and task execution in professional environments.

\paragraph{Metrics.}
We assess performance and operational cost using a multi-dimensional metric suite. The primary behavioral metric is \textbf{Accuracy}, which measures the success rate of task completion. To quantify operational overhead, we report \textbf{Input Tokens}, \textbf{Output Tokens}, and their sum, \textbf{Total Tokens}. Finally, we define \textbf{Efficiency} as the accuracy achieved per million total tokens consumed, formalized as $\text{Efficiency} = \text{Accuracy} / \text{Total Tokens (M)}$.



\subsubsection{Results}
\label{sec:exp_task_completion_results}

\textbf{GA consistently achieves state-of-the-art or highly competitive task completion rates.}
As detailed in Table~\ref{tab:dimension1_task_completion}, GA's robust performance is consistent across all three benchmarks. In the Claude-based configurations, GA achieves 100\% accuracy on both SOP-Bench and Lifelong AgentBench. This perfectly matches the highest baseline performance on SOP-Bench and strictly outperforms all competitors on Lifelong AgentBench. This superiority extends to the RealFin-Benchmark, where GA attains the highest overall accuracy of 65\%, comfortably exceeding Claude Code (60\% with \textit{Claude Opus 4.6} and 55\% with \textit{Claude Sonnet 4.6}), CodeX (60\%), and OpenClaw (35\%).

\textbf{GA significantly reduces token consumption, particularly in input contexts.}
Beyond absolute task success, GA demonstrates marked token efficiency. On Lifelong AgentBench, GA requires only 222k input tokens, representing a drastic reduction compared to Claude Code (800k) and OpenClaw (1.43M), while seamlessly maintaining its flawless 100\% accuracy. On SOP-Bench, GA consumes 2.02M input tokens---substantially less than OpenClaw (2.60M). Although Claude Code uses nominally fewer tokens (1.23M) on this specific benchmark, it does so at the steep cost of task completion (as analyzed below). Similar cost-saving patterns are consistently observed in the RealFin-Benchmark.

\textbf{GA achieves the strongest overall performance in task completion and efficiency across the main benchmarks.}
Within each benchmark, GA achieves the best combined result in terms of task completion and token efficiency. On Lifelong AgentBench, GA reaches 100\% accuracy with an efficiency ratio of 4.15; on RealFin-benchmark, it attains 65\% accuracy with an efficiency ratio of 5.70. On SOP-Bench, although Claude Code reaches a higher efficiency ratio of 0.68, its accuracy drops to 85\%, whereas GA maintains 100\% accuracy with an efficiency ratio of 0.48.

\subsection{Tool-use efficiency}
\label{sec:dimension4_tool_use_efficiency}

This section evaluates the impact of tool space design on an agent's operational efficiency and problem-solving capability. The primary objective is to investigate whether a minimal, atomic-tool abstraction can effectively handle complex workflows that typically rely on a large inventory of specialized tools. By analyzing tool usage distributions, interaction overhead, and task success rates, this experiment aims to verify if condensing the action space into a small set of core composable functions can mitigate prompt bloat and reduce interaction cycles without compromising the agent's functional versatility.

\subsubsection{Setup}
\label{sec:exp_tooluse_setup}

\paragraph{Baseline.}
All compared systems use the same backbone model, \textit{Claude Sonnet 4.6}. We evaluate three representative agents: (1) \textbf{GA},  (2) \textbf{Claude Code}, (3) \textbf{OpenClaw}.

\paragraph{Tool.}
Table~\ref{tab:tool_inventory_overview} first gives a source-level inventory overview. Claude Code exposes 53 built-in tools at the source level (20 base tools and 33 conditional tools), while OpenClaw provides 18 source-level tool factories, with the runtime-visible set further varying by plugin injection and environment configuration. GA does not attempt to reproduce these full inventories one by one. Instead, it keeps only a small set of core atomic tools that cover the capabilities most frequently required in our evaluation tasks, and each of these capabilities has a clear counterpart in both Claude Code and OpenClaw. Appendix~\ref{app:atomic_tool_alignment} provides both a capability-level alignment (Table~\ref{tab:appendix_tool_alignment}) and representative task-level substitution examples (Table~\ref{tab:appendix_tool_replacement_examples}).

\paragraph{Benchmark.}
The benchmark contains two complementary task groups. (1) \textbf{Simple tool-generalization tasks}, which test whether baseline specialized tool capabilities can be reproduced through GA's atomic-tool composition. (2) \textbf{Long-horizon complex tasks}, which test whether the same minimal tool design remains effective on realistic multi-step workflows.

\paragraph{Metrics.}
We report \textbf{Success}, \textbf{Total Tokens}, \textbf{Time}, \textbf{Requests}, and \textbf{Tool Calls}. \textbf{Success} denotes the fraction of tasks completed successfully in the evaluated task set. The remaining metrics measure execution cost and interaction overhead: \textbf{Total Tokens} is the total token consumption, \textbf{Time} is the average runtime in seconds, \textbf{Requests} is the average number of model requests, and \textbf{Tool Calls} is the average number of tool invocations.

\subsubsection{Atomic Tool Generalization}
\label{sec:exp_atomic_tool_generalization}

\begin{table*}[t]
\centering
\caption{\textbf{Results on long-horizon complex tasks.} The five tasks cover document generation (PDF/PPT creation), SQL copilot query generation, experiment analysis report writing, procurement decision-making with web retrieval, and research paper reproduction feasibility analysis. This table reports the average outcomes over the long-horizon task set rather than expanding into per-task detail.}
\def\arraystretch{0.99}
\setlength{\tabcolsep}{0.42em}
\resizebox{0.8\linewidth}{!}{
\begin{tabular}{lcccccc}
\toprule
\textbf{Agent} & \textbf{\#Tasks} & \textbf{Success} & \textbf{Total Tokens} & \textbf{Time (s)} & \textbf{Requests} & \textbf{Tool Calls} \\
\midrule
Claude Code & 5 & 100.0\% & 537,413 & 320.8 & 32.6 & 22.6 \\
GA & 5 & 100.0\% & 188,829 & 220.8 & 11.0 & 12.8 \\
OpenClaw & 5 & 80.0\% & 633,101 & 183.1 & 15.0 & 16.6 \\
\bottomrule
\end{tabular}
}
\label{tab:tooluse_long_horizon}
\end{table*}

\textbf{GA can solve complex long-horizon tasks through compositions of atomic tools rather than through a richer specialized tool inventory.} In Table~\ref{tab:tooluse_long_horizon}, GA reaches 100\% success on the five long-horizon tasks, matching Claude Code and outperforming OpenClaw, which falls to 80\%. This shows that a minimal tool set does not reduce the range of tasks that can be solved when the tools are composable in the right way. Appendix~\ref{app:atomic_tool_alignment}, especially Table~\ref{tab:appendix_tool_replacement_examples}, provides concrete substitution examples showing how several specialized baseline capabilities can be reconstructed through short atomic-tool compositions.

\textbf{GA reduces token overhead while preserving task performance.} Although it matches Claude Code on success, GA uses only 188,829 accounted tokens, which is 35.1\% of Claude Code's 537,413 and 29.8\% of OpenClaw's 633,101. The same pattern appears in the interaction statistics: GA reduces requests from 32.6 to 11.0 and tool calls from 22.6 to 12.8 relative to Claude Code, while OpenClaw remains both less stable and more expensive. These results indicate that GA's advantage comes from lower context and tool-process overhead, not from sacrificing success rate.

\subsubsection{Tool Usage Distribution}
\label{sec:exp_tool_usage_distribution}

\begin{figure*}[t]
\centering
\includegraphics[width=0.9\linewidth]{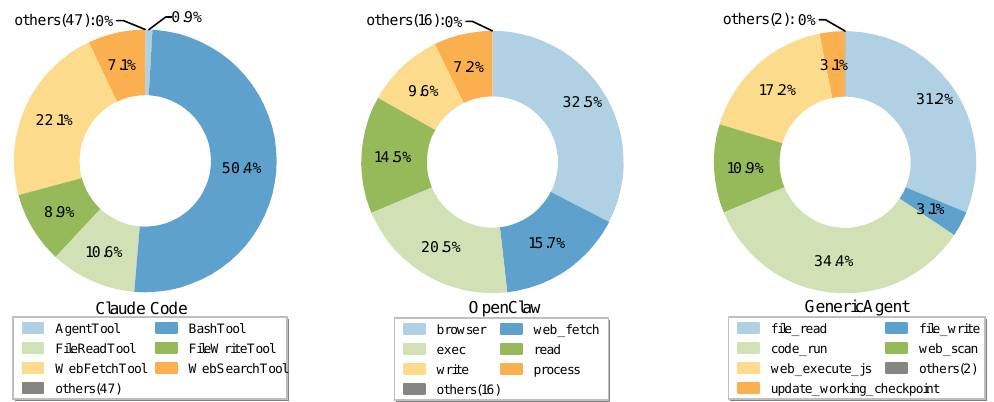}
\caption{\textbf{Tool usage distribution across Claude Code, OpenClaw, and GA.} Each donut chart summarizes the proportion of tool calls attributed to each major tool category in the corresponding system.}
\label{fig:tooluse_distribution}
\end{figure*}

\textbf{Tool usage is highly concentrated in a few high-frequency tools, while the remaining long-tail tools are invoked only rarely.} As shown in Figure~\ref{fig:tooluse_distribution}, although Claude Code and OpenClaw expose dozens of built-in tools or tool factories, their actual execution traces are still dominated by a small subset of tools. In Claude Code, \texttt{AgentTool} alone accounts for 50.4\% of calls, followed by \texttt{WebFetchTool} (22.1\%), \texttt{FileReadTool} (10.6\%), and \texttt{FileWriteTool} (8.9\%), while many other tools appear only in the tail. OpenClaw shows a similar concentration pattern. This means that a substantial number of low-frequency tools still occupy prompt context and enlarge the action space even though they contribute little to actual execution. By comparison, GA makes the active tool loop explicit through a compact atomic-tool set centered on \texttt{code\_run}, \texttt{file\_read}, \texttt{web\_execute\_js}, and \texttt{web\_scan}, which is consistent with its lower request count, lower tool-call count, and lower token cost in Table~\ref{tab:tooluse_long_horizon}.

\subsection{Memory System Effectiveness}
\label{sec:exp_memory}


This section evaluates the architectural design and operational impact of GA's hierarchical memory system. The primary objective is to investigate whether the framework can effectively accumulate, distill, and retrieve historical experience without incurring the prohibitive cost of context explosion. By systematically analyzing performance dynamics across repeated tasks, memory compression strategies, factual retention mechanisms, and prompt growth over prolonged usage, this experiment aims to verify that high-density, filtered memory significantly enhances continuous learning while strictly maintaining an optimal context boundary.

\subsubsection{Continuous Efficiency Improvement}
\label{sec:exp_continuous_efficiency}

\paragraph{Setup.}
To evaluate whether the Condensed Memory framework enables GA to improve through continuous real-world use, we assess the performance trajectories of different agents across repeated runs. We compare \textbf{CodeX}, \textbf{Claude Code}, \textbf{OpenClaw}, and \textbf{GA}, all utilizing \textit{GPT-5.4} as the backbone model. Each experimental run involves downloading five distinct datasets from HuggingFace. To strictly avoid contamination from residual conversational context, every task is executed within a completely fresh session.


\begin{figure*}[t]
\centering
\includegraphics[width=0.9\linewidth]{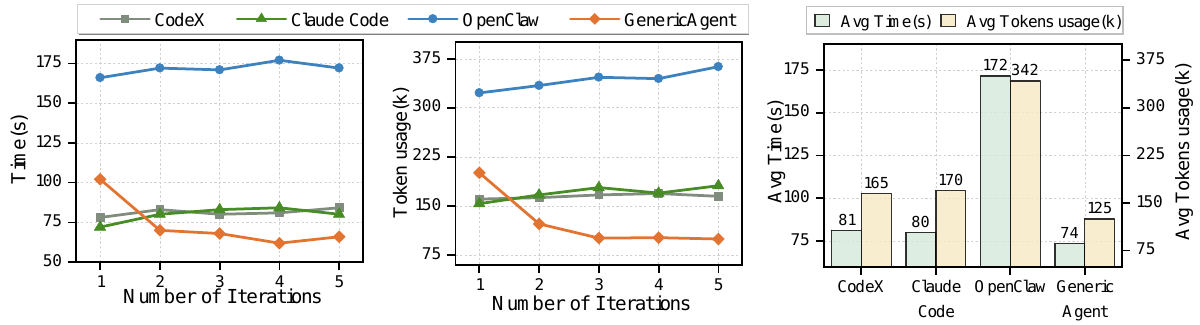}
\caption{Operation time and token consumption across five repeated runs for different agents. While the baseline agents remain relatively stable over time, GA shows a clear efficiency convergence pattern, with both runtime and token cost decreasing substantially as task experience is accumulated and distilled into reusable memory.}
\label{fig:memory}
\end{figure*}

\paragraph{Results.}
\textbf{GA continuously improves operational efficiency through repeated use by successfully converting raw task experience into reusable memory.}
As shown in Figure~\ref{fig:memory}, CodeX, Claude Code, and OpenClaw remained largely stable across runs, whereas GA improved markedly, with operation time decreasing from 102 seconds in the first run to approximately 66 seconds in later runs, and token consumption dropping from 200,439 to 100,000.
This steady reduction indicates that GA does not simply accumulate history, but converts task experience into reusable L3 SOPs that reduce repeated understanding, unnecessary reasoning, and decision overhead in later runs. 
More importantly, the gain comes not from storing more content, but from preserving high-value information that the model does not already know and that directly affects behavior. 
Overall, the results show that GA becomes more efficient through continued use, which is precisely the intended effect of its memory system.

\subsubsection{Effectiveness of Condensed memory}
\label{sec:exp_condensed_memory}

\paragraph{Setup.}
To examine whether retaining only decision-critical rules yields better performance at a lower contextual cost, we compare four distinct memory configurations under the GA framework (\textit{GPT-5.4}):
(1) \textbf{No-Memory}: Operates solely on task inputs and tool descriptions without external memory;
(2) \textbf{Full-Memory}: Injects the complete, unedited SOP as external knowledge;
(3) \textbf{Redundant-Memory}: Extends Condensed Memory with background descriptions, definitions, and weakly relevant information;
(4) \textbf{Condensed Memory}: Retains only high-density, action-guiding rules.
The evaluation is conducted on the dangerous goods subset of SOP-Bench~\citep{sopbench}.
We use \textbf{Task Success Rate (TSR)} to measure behavioral performance and Memory Size to measure contextual cost, and we further compare the performance gain achieved per unit of token cost across different memory settings.


\begin{table}[t]
\centering
\caption{Memory ablation on SOP-Bench dangerous\_goods. Condensed memory achieves the best cost-effectiveness.}
\label{tab:memory_ablation}
\begin{tabular}{lcc}
\toprule
\textbf{Configuration} & \textbf{Memory Size (tokens)} & \textbf{TSR (\%)} \\
\midrule
No-Memory & 0 & 13.87 \\
Full-Memory & 575 & 52.44 \\
Redundant-Memory & 288 & 66.48 \\
Condensed memory & 165 & \textbf{66.48} \\
\bottomrule
\end{tabular}
\end{table}

\paragraph{Results.}
\noindent\textbf{Filtered, high-density memory provides optimal behavioral guidance while drastically minimizing contextual burden.}
As shown in Table~\ref{tab:memory_ablation}, No-Memory performed substantially worse than all memory based settings, which shows that external procedural memory is necessary for this task. 
Condensed memory achieved the same highest TSR as Redundant-Memory and outperformed Full-Memory while using far fewer tokens, which indicates that filtered high density memory is more effective than the original full SOP. 
Redundant-Memory provided no gain over Condensed memory despite using more tokens, which suggests that background descriptions, definitions, and explanatory text add contextual burden without contributing additional behavioral value. 
Overall, this experiment shows that GA benefits most from memory that stores only the small set of task critical information the model does not already know.

\subsubsection{Long-Term Fact Retention}
\label{sec:exp_longterm_fact}

\paragraph{Setup.}
This experiment examines whether GA has long term factual memory and whether it can remain competitive on factual recall and multi hop reasoning tasks without relying on an additional embedding model or a vector database. 
The compared methods include embedding based approaches, namely Mem0~\citep{mem0} and A-MEM, and non embedding approaches, namely OpenClaw and GA. 
All methods use \textit{GPT-5.4}, and for the embedding based methods we adopt \textit{text-embedding-3-small}) as the embedding model. 
The evaluation is conducted on the first subset of LoCoMo~\citep{locomo}, with Category 5 summary tasks removed to avoid interference from summarization ability.
We use F1 to measure the correctness and completeness of factual content, and BLEU-1 to measure the lexical similarity between the generated answer and the reference answer.

\begin{table}[t]
\centering
\caption{Long-term factual memory evaluation on LoCoMo. GA outperforms embedding-based systems without using external retrieval.}
\label{tab:long_term_memory}
\begin{tabular}{lcccccccc}
\toprule
& \multicolumn{2}{c}{\textbf{Multi-Hop}} & \multicolumn{2}{c}{\textbf{Temporal}} & \multicolumn{2}{c}{\textbf{Open-Domain}} & \multicolumn{2}{c}{\textbf{Single-Hop}} \\
\cmidrule(lr){2-3} \cmidrule(lr){4-5} \cmidrule(lr){6-7} \cmidrule(lr){8-9}
\textbf{System} & \textbf{F1} & \textbf{BLEU} & \textbf{F1} & \textbf{BLEU} & \textbf{F1} & \textbf{BLEU} & \textbf{F1} & \textbf{BLEU} \\
\midrule
Mem0 & 39.32 & 28.43 & 50.03 & 43.33 & 18.32 & 13.80 & 40.32 & 32.33 \\
A-MEM & 29.03 & 20.48 & 46.83 & 38.84 & 13.11 & 12.94 & 44.68 & 37.01 \\
OpenClaw & 21.43 & 18.41 & 22.56 & 20.43 & 9.56 & 9.03 & 23.44 & 24.21 \\
GA & \textbf{43.33} & \textbf{39.96} & \textbf{52.23} & \textbf{51.11} & \textbf{20.41} & \textbf{15.31} & \textbf{45.69} & \textbf{40.66} \\
\bottomrule
\end{tabular}
\end{table}

\paragraph{Results.}
\noindent\textbf{Precise memory organization enables GA to exceed the factual retention capabilities of dedicated vector-based retrieval systems.}
As shown in Table~\ref{tab:long_term_memory}, GA achieved the best F1 and BLEU-1 scores across all four task categories, which indicates that its memory mechanism supports both accurate factual recall and stable answer generation. 
Its advantage was especially clear on Multi-Hop and Temporal tasks, which suggests that GA can not only retain long term facts but also use them effectively in reasoning across time and across fact chains. 
Even in Open-Domain, which was the most difficult category for all methods, GA still achieved the highest score, and this shows that its memory organization remains effective even when structural cues are weak. 
Overall, these results show that long term factual memory can be achieved through precise memory organization and filtering, and that GA does not need an additional embedding model or a vector database to build this capability.

\subsubsection{Context Explosion Prevention}
\label{sec:exp_context_explosion}
\paragraph{Setup.}
To evaluate whether GA can prevent memory explosion under long term use and continuous skill expansion, we compare \textbf{GA}, \textbf{Claude Code}, \textbf{CodeX}, and \textbf{OpenClaw} after installing the same set of 20 skills for each agent and using them intensively over time. 
We then issue a minimal request ``Hello'' and measure the Full prompt length of each agent to assess how well its memory system controls context growth after prolonged use.

\begin{table}[t]
\centering
\caption{Full prompt length after installing 20 skills and intensive usage, measured on a minimal input. GA prevents context explosion.}
\label{tab:context_explosion}
\begin{tabular}{lc}
\toprule
\textbf{System} & \textbf{Full Prompt Length (tokens)} \\
\midrule
Claude Code & 22,821 \\
CodeX & 23,932 \\
OpenClaw & 43,321 \\
GA & \textbf{2,298} \\
\bottomrule
\end{tabular}
\end{table}

\paragraph{Results.}
\noindent\textbf{Hierarchical retrieval strictly isolates idle memory from the active prompt, eliminating the risk of context explosion.}
As shown in Table~\ref{tab:context_explosion}, the Full prompt length of GA is far lower than that of Claude Code, CodeX, and OpenClaw, which shows that GA can keep context growth under control even after long term use and continuous skill expansion. 
This advantage comes from its hierarchical memory management, where memory is not directly injected into the context but is retrieved only when the model actually needs it. 
Overall, the results show that GA can preserve long term memory capability without causing uncontrolled context expansion.

\subsection{Self-Evolution Capability}
\label{sec:self_evolution_capability}

This section evaluates the framework's capacity for self-evolution, specifically its ability to distill and reuse experiential knowledge over time. The primary objective is to investigate whether an agent can systematically transition from high-entropy, trial-and-error exploration to a deterministic, low-cost execution regime. By tracking performance trajectories across repeated interactions and varying task complexities, this experiment aims to verify if the reflection-driven pipeline effectively compresses historical interaction trajectories into reusable SOPs and executable code, thereby permanently eliminating redundant reasoning cycles and preventing stagnation loops in long-horizon deployments.




\subsubsection{Setup}
\label{sec:exp_selfevolution_setup}

\paragraph{Baseline.}
To rigorously evaluate the cross-task generalization of the self-evolution mechanism, we establish a direct comparison between \textbf{GA} and \textbf{OpenClaw}. To ensure a fair baseline and isolate the impact of system architecture from model capability, both agents utilize \textit{Claude Opus 4.6} as their underlying backbone model for the comparative benchmark.

\paragraph{Evaluation Setting.}
We evaluate GA's self-evolution process across two complementary experimental settings:
(1) \textbf{Longitudinal Study}: A nine-round sequential evaluation where GA repeatedly performs GitHub research tasks. Each round operates on a new task instance, allowing us to trace the full evolutionary trajectory and observe how accumulated experience translates into progressive efficiency gains on structurally similar workflows.
(2) \textbf{Cross-Task Benchmark}: An eight-task web benchmark designed to rigorously test whether the efficiency convergence observed in the longitudinal study generalizes across diverse task types and environments when compared against the OpenClaw baseline.
The primary metric for evaluating self-evolution efficiency is \textbf{Total Tokens} (token consumption), which is tracked across repeated executions to quantify the reduction in contextual and reasoning overhead.

\paragraph{Evolution Stages.}
To systematically analyze the evolution process, we categorize GA's capability transformations into three distinct stages:
(1) \textbf{Stage 1 (Natural-language execution)}: The agent solves the task primarily through in-context reasoning, exploratory tool use, and trial-and-error interaction.
(2) \textbf{Stage 2 (SOP distillation)}: Accumulated experiential memory is compressed into a structured, textual SOP, filtering out exploratory missteps.
(3) \textbf{Stage 3 (Code-based execution)}: The verified textual workflow is further crystallized into executable logic.
Crucially, the transitions between these stages are triggered autonomously by GA's memory management mechanisms without manual intervention, serving as a core architectural advantage.

\subsubsection{Iteration-Wise Efficiency Trajectory}
\label{sec:exp_iteration_trajectory}

\begin{table*}[t]
\centering
\caption{\textbf{Nine-round evolution trajectory on the LangChain GitHub research task.} The experiment starts from scratch in Round \#1, iteratively refines a textual SOP in Rounds \#2--\#5, and enters the codified execution regime in Rounds \#6--\#9.}
\def\arraystretch{0.99}
\setlength{\tabcolsep}{0.42em}
\resizebox{1.0\linewidth}{!}{
\begin{tabular}{l l c c c c c c c}
\toprule
\textbf{Round} & \textbf{Stage} & \textbf{Time} & \textbf{LLM Calls} & \textbf{Input} & \textbf{Output} & \textbf{Cache Create} & \textbf{Cache Read} & \textbf{Total} \\
\midrule

\#1 & Initial run & 7m30s & 32 & 15,581 & 7,647 & 15,600 & 183,375 & 222,203 \\
\#2 & SOP optimization & 4m19s & 12 & 5,045 & 4,860 & 4,553 & 51,883 & 66,341 \\
\#3 & SOP optimization & 2m53s & 8 & 2,538 & 4,598 & 3,155 & 39,534 & 49,825 \\
\#4 & SOP optimization & 2m29s & 9 & 3,265 & 3,682 & 3,435 & 41,376 & 51,758 \\
\#5 & SOP optimization & 2m50s & 7 & 2,568 & 3,276 & 2,424 & 27,268 & 35,536 \\
\#6 & Codified SOP & 2m24s & 6 & 1,855 & 1,064 & 1,930 & 20,913 & 25,762 \\
\#7 & Codified SOP & 1m41s & 5 & 1,931 & 1,126 & 1,708 & 18,249 & 23,014 \\
\#8 & Codified SOP & 1m35s & 5 & 1,884 & 990 & 1,586 & 18,229 & 22,689 \\
\#9 & Codified SOP & 1m38s & 5 & 1,323 & 994 & 1,659 & 19,034 & 23,010 \\

\bottomrule
\end{tabular}
}
\label{tab:self_evolution_rounds}
\end{table*}

\textbf{The nine-round trajectory shows a phase transition from high-entropy exploration to a stable low-cost execution regime.} 
As shown in Table~\ref{tab:self_evolution_rounds}, relative to Round \#1, the final execution in Round \#9 reduces runtime from 7m30s to 1m38s, a reduction of 78.2\%, reduces LLM calls from 32 to 5, a reduction of 84.4\%, and reduces total tokens from 222,203 to 23,010, a reduction of 89.6\%. 
The key pattern is not just the large first-to-last gap, but the structure of the transition itself. 
During the textual SOP stage, token cost decreases overall from 66k to 36k, despite minor fluctuations, which indicates that SOP compression removes most redundant exploration while still incurring a limited adaptation cost when the stored procedure must be aligned with newly encountered edge cases.
Once the workflow is codified, the system enters a narrow convergence band of roughly 23k$\pm$1k tokens across Rounds \#6--\#9, indicating that execution has become predictable rather than exploratory.

\textbf{Most of the efficiency gain comes from eliminating repeated reasoning cycles rather than merely shortening individual responses.} The dominant change is the collapse in call count, which removes entire understand--reason--generate loops from the execution trace. This is consistent with the token composition: cache-read tokens fall from 183,375 in Round \#1 to 19,034 in Round \#9, and input tokens fall from 15,581 to 1,323, showing that the agent no longer rebuilds large amounts of context at runtime. Per-call token load also drops substantially over the course of evolution, from 6,944 tok/call in Round \#1 to about 4.5k--4.6k tok/call in Rounds \#7--\#9, but this effect is secondary to the structural reduction in the number of calls. The larger implication is that textual SOPs already compress the search space, while code-level execution removes the remaining interpretation overhead and turns the learned workflow into an executable module.

\subsubsection{Cross-Task Efficiency Gains}
\label{sec:exp_crosstask_efficiency}
\begin{figure*}[t]
\centering
\includegraphics[width=0.99\linewidth]{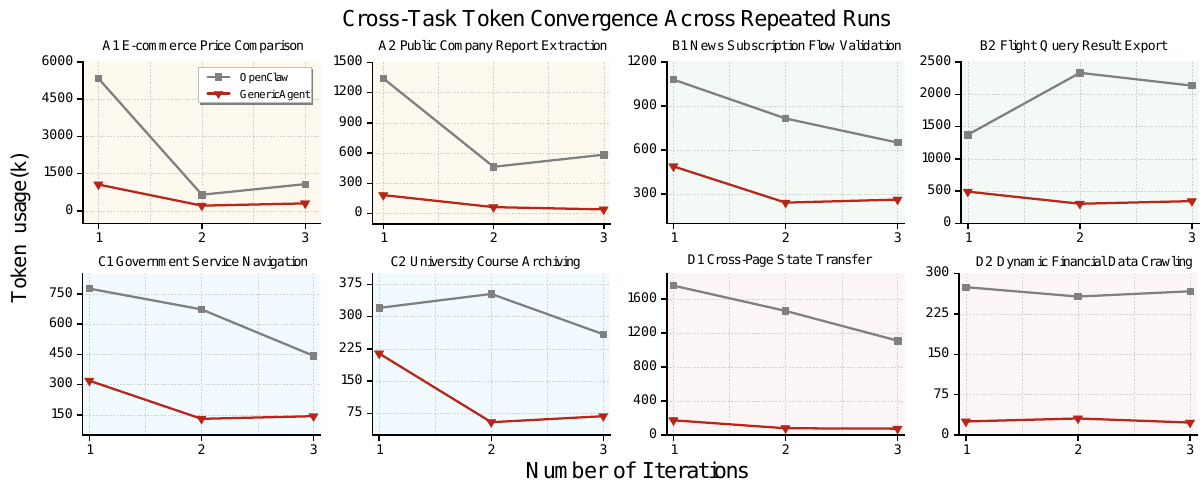}
\caption{\textbf{Cross-task token convergence across repeated runs for OpenClaw and GA.} Each panel corresponds to one benchmark task and plots total token consumption over the first, second, and third executions.}
\label{fig:cross_task_convergence}
\end{figure*}
\textbf{GA's self-evolving memory mechanism consistently improves efficiency across tasks rather than only helping on a single hand-optimized workflow.} As shown in Figure~\ref{fig:cross_task_convergence}, on all eight benchmark tasks, later GA executions consume fewer tokens than the first execution, with savings ranging from 61.0\% to 92.4\% and an overall reduction of 79.3\%. This consistency matters more than any individual best case: it shows that SOP self-evolution is not just memorizing one task, but systematically converting prior successful trajectories into reusable execution shortcuts. The strongest gains appear in long-horizon tasks with state transfer and recovery (Category D, 92.0\%), where repeated failure handling and path search dominate the initial execution cost. In those cases, SOP acts as path compression: once the verified route is stored, the agent can bypass most exploratory branches entirely.

\textbf{GA exhibits a clear transition from cold start to rapid convergence across repeated executions of the same task.} Across the three runs of each benchmark task, GA shows the same internal trend: the first run has substantially higher token cost, while the second and third runs quickly converge to a stable low-cost regime. This indicates that even with SOP memory, the first execution still incurs an SOP adaptation cost, because the agent must map generalized SOP knowledge onto the concrete page structure and interaction flow of the current task instance. However, this adaptation happens only once. After the successful path is aligned with the environment, later executions can directly reuse that result and rapidly drop to a stable token level. By contrast, OpenClaw shows no comparable convergence pattern across its three runs. On B2, for example, token usage changes from 1,370k to 2,330k to 2,130k, which means the system keeps re-exploring rather than reusing prior experience. This distinction is important for deployment: GA pays a one-time adaptation cost on a new task, but its marginal cost quickly approaches zero, whereas the baseline remains expensive on every execution.

\textbf{The benefit of SOP memory tends to increase with task complexity.} When the tasks are ordered by OpenClaw's average token consumption as a proxy for intrinsic task complexity, a positive relationship emerges between complexity and saving rate. High-complexity tasks with OC averages above 1,000k tokens, such as A1, B2, and D1, achieve an average saving of 83.5\%, while mid-complexity tasks average 72.5\%. Although D2 is low in absolute token cost, its 90.1\% saving rate is consistent with the same mechanism because it still contains long-horizon dependency structure. The broader implication is that SOP self-evolution becomes more valuable as tasks require more multi-step reasoning, navigation, and recovery from branching execution paths: in simple tasks it is helpful, but in complex tasks it becomes structurally decisive.

\subsection{Web Browsing Capability}
\label{sec:exp_browsing}

This section evaluates the agent's capability to navigate, search, and reason within open-ended web environments. The primary objective is to use the unstructured, noise-heavy nature of the web as a comprehensive stress test for the system's context management and multi-step execution. By requiring the agent to interact with dynamic web pages, filter irrelevant interface elements, and conduct multi-hop retrieval, this experiment aims to verify whether GA's principles of high-density context and hierarchical memory remain robust in real-world scenarios where prompt explosion from raw HTML or DOM structures is a typical failure mode.

\subsubsection{Setup}
\label{sec:exp_browsing_setup}

\paragraph{Baseline.}
We compare \textbf{GA} against \textbf{OpenClaw} to contrast their architectural approaches in handling highly dynamic environments. To ensure a strictly fair comparison focused on system design rather than model capability, both GA and OpenClaw utilize \textit{Claude 4.6 Opus} as the underlying backbone model.

\paragraph{Benchmark.}
We evaluate web browsing proficiency across three diverse benchmarks, ranging from atomic interactions to open-ended real-world workflows:
(1) \textbf{WebCanvas}~\citep{webcanvas}: A benchmark measuring fundamental browser interactions, including navigation, element clicking, filtering, and localized information extraction. We randomly sample 12 tasks for evaluation.
(2) \textbf{BrowseComp-ZH}~\citep{browsecomp_zh}: A benchmark specifically designed to measure multi-hop web search and chain-based reasoning within the Chinese web ecosystem. We randomly sample 10 tasks for evaluation.
(3) \textbf{Custom Tasks}: A curated set of 22 real-world tasks encompassing academic platforms, social media, content sites, utility websites, e-commerce platforms, and public retrieval services.

\paragraph{Metrics.}
Performance is evaluated using a normalized \textbf{Score (0-1)}. The scoring protocols are tailored to each benchmark: automatic evaluation for WebCanvas, LLM-as-a-judge for BrowseComp-ZH, and a rigorous human-in-the-loop review assisted by GPT-5.4 for Custom Tasks. To quantify operational overhead in web environments, we also report the \textbf{Avg. Tokens (M)} consumed per task.

\subsubsection{Results}
\label{sec:exp_browsing_results}

\textbf{GA consistently outperforms the baseline in web navigation and reasoning while operating at a fraction of the token cost.}
As shown in Table~\ref{tab:web_browsing_results}, GA achieves superior scores across all three benchmarks compared to OpenClaw. On the fundamental WebCanvas benchmark, GA secures a score of 0.834 using only 0.18M tokens, whereas OpenClaw trails at 0.722 while consuming 0.71M tokens. This advantage extends to open-ended, real-world interactions: on Custom Tasks, GA scores 0.577 against OpenClaw's 0.500, while utilizing a mere 0.26M tokens compared to OpenClaw's 0.76M. 

\begin{table}[t]
\centering
\small
\caption{Web browsing evaluation results across three benchmarks. GA and OpenClaw both use \textit{Claude Opus 4.6} as the backbone model.}
\begin{tabular}{lcccccc}
\toprule
\multirow{2}{*}{Benchmark} & \multirow{2}{*}{Tasks} & \multirow{2}{*}{Evaluation} & \multicolumn{2}{c}{Score (0--1)} & \multicolumn{2}{c}{Avg. Tokens (M)} \\
\cmidrule(lr){4-5} \cmidrule(lr){6-7}
 &  &  & GA & OpenClaw & GA & OpenClaw \\
\midrule
WebCanvas & 12 & Automatic & \textbf{0.834} & 0.72 & \textbf{0.18} & 0.71 \\
BrowseComp-ZH & 10 & LLM Judge & \textbf{0.60} & 0.20 & \textbf{0.47} & 1.31 \\
Custom Tasks & 22 & Human + LLM & \textbf{0.577} & 0.50 & \textbf{0.26} & 0.76 \\
\bottomrule
\end{tabular}
\label{tab:web_browsing_results}
\end{table}

\textbf{GA's multi-stage compression pipeline provides an advantage in long-horizon, multi-hop web search tasks.}
The performance gap becomes exceptionally pronounced on the BrowseComp-ZH benchmark, which demands deep reasoning chains. Here, GA achieves an impressive 0.600 accuracy, tripling OpenClaw's score of 0.200, while consuming roughly one-third of the tokens (0.47M vs. 1.31M). Without active memory filtering, baseline systems like OpenClaw are easily overwhelmed by raw HTML and irrelevant DOM elements, leading to state confusion and context overflow. Conversely, GA preserves context density by systematically compressing web observations, enabling the model to sustain deep, multi-hop logical reasoning without losing track of the initial objective.

\textbf{The framework demonstrates a structural 2.9x to 3.9x reduction in token consumption, making it highly practical for real-world automation.}
Across diverse web tasks, GA exhibits remarkable and consistent token efficiency. This substantial reduction in token consumption is not merely an economic benefit; it directly translates to lower latency and higher reliability, which are critical constraints for deployable web agents. This combination of robust effectiveness and rigorous context control validates GA as a highly practical architecture for real-world, open-ended web automation scenarios. A detailed qualitative visualization of GA's web execution trajectory is provided in Appendix~\ref{app:web_browsing_visualization}.



%% file: tables/main_result.tex
\begin{table*}[t]
\centering
\caption{\textbf{Task completion rate and token efficiency across the main agent benchmarks and RealFin-benchmark.} Each block reports the original metrics used in the corresponding benchmark. Since the efficiency ratio is normalized with different token scales across benchmarks, its absolute values should only be compared within the same benchmark block.}
\def\arraystretch{0.99}
\setlength{\tabcolsep}{0.45em}
\resizebox{1.0\linewidth}{!}{
\begin{tabular}{llccccc}
\toprule
\textbf{Agent} & \textbf{Model} & \textbf{Accuracy} & \textbf{Input Tokens} & \textbf{Output Tokens} & \textbf{Total Tokens} & \textbf{Efficiency} \\
\midrule
\rowcolor{bgcolor}
\multicolumn{7}{c}{\textbf{SOP-Bench}} \\
GA & Claude Sonnet 4.6 & 100\% & 2.02M & 53k & 2.08M & 0.48 \\
OpenClaw & Claude Sonnet 4.6 & 100\% & 2.60M & 40k & 2.64M & 0.38 \\
Claude Code & Claude Sonnet 4.6 & 85\% & 1.23M & 23k & 1.25M & 0.68 \\
GA & Minimax M2.7 & 90\% & 893k & 32k & 924k & 0.97 \\
OpenClaw & Minimax M2.7 & 95\% & 2.91M & 46k & 2.96M & 0.32 \\
\midrule
\rowcolor{bgcolor}
\multicolumn{7}{c}{\textbf{Lifelong AgentBench}} \\
GA & Claude Sonnet 4.6 & 100\% & 222k & 20k & 241k & 4.15 \\
OpenClaw & Claude Sonnet 4.6 & 70\% & 1.43M & 21k & 1.45M & 0.48 \\
Claude Code & Claude Sonnet 4.6 & 75\% & 800k & 14k & 814k & 0.92 \\
GA & Minimax M2.7 & 90\% & 400k & 23k & 423k & 2.12 \\
OpenClaw & Minimax M2.7 & 70\% & 1.20M & 17k & 1.22M & 0.57 \\
\midrule
\rowcolor{bgcolor}
\multicolumn{7}{c}{\textbf{RealFin-benchmark}} \\
GA & Claude Sonnet 4.6 & 65\% & 102k & 12k & 114k & 5.70 \\
Claude Code & Claude Opus 4.6 & 60\% & 290k & 17k & 307k & 1.95 \\
Claude Code & Claude Sonnet 4.6 & 55\% & 226k & 12k & 238k & 2.31 \\
OpenClaw & Claude Sonnet 4.6 & 35\% & 249k & 2k & 251k & 1.39 \\
Codex & GPT-5.4 & 60\% & 838k & 54k & 892k & 0.67 \\
\bottomrule
\end{tabular}
}
\label{tab:dimension1_task_completion}
\end{table*}

%% file: tables/tool_inventory_overview.tex
\setlength\tabcolsep{5.2pt}
\begin{table*}[t]
\centering
\def\arraystretch{1.02}
\setlength{\tabcolsep}{0.45em}
\caption{\textbf{Source-level tool inventory overview of Claude Code, OpenClaw, and GA.} We list representative native tools for Claude Code and OpenClaw and show the full atomic tool set of GA. The counts refer to source-level built-in tools/tool factories rather than the exact runtime-visible tool set, which may vary with configuration, feature flags, permissions, and plugin-based dynamic injection.}
\resizebox{0.98\linewidth}{!}{
\begin{tabular}{p{0.33\linewidth} p{0.33\linewidth} p{0.24\linewidth}}
\toprule
\textbf{Claude Code} & \textbf{OpenClaw} & \textbf{GA} \\
\midrule
\makecell[tl]{\texttt{AgentTool}\\
\texttt{TaskOutputTool}\\
\texttt{BashTool}\\
\texttt{GlobTool}\\
\texttt{GrepTool}\\
\texttt{FileReadTool}\\
\texttt{FileEditTool}\\
\texttt{FileWriteTool}\\
\texttt{NotebookEditTool}\\
\texttt{WebFetchTool}\\
\ldots\\
\textit{53 source-level built-in tools:}\\
\textit{20 base + 33 conditional.}}
&
\makecell[tl]{\texttt{browser}\\
\texttt{canvas}\\
\texttt{nodes}\\
\texttt{cron}\\
\texttt{message}\\
\texttt{tts}\\
\texttt{gateway}\\
\texttt{agents-list}\\
\texttt{sessions-list}\\
\texttt{sessions-history}\\
\ldots\\
\textit{18 source-level tool factories;}\\
\textit{runtime may include plugins.}}
&
\makecell[tl]{\texttt{file\_read}\\
\texttt{file\_write}\\
\texttt{file\_patch}\\
\texttt{code\_run}\\
\texttt{web\_scan}\\
\texttt{web\_execute\_js}\\
\texttt{ask\_user}\\
\texttt{update\_working\_}\\
\texttt{checkpoint}\\
\texttt{start\_long\_}\\
\texttt{term\_update}\\
\textit{only 9 atomic tools}}\\
\bottomrule
\end{tabular}
}
\label{tab:tool_inventory_overview}
\end{table*}

%% file: section/6discussion.tex
We highlight key findings from the development of GenericAgent that 
we believe are broadly relevant to the design of long-horizon LLM 
agent systems. These observations are distilled from practice rather 
than derived from controlled ablations alone, and we hope they will 
inform and inspire future research.

\textbf{Context information density is a structural constraint for all 
LLM-based agent systems.} Prior work has established that LLM reasoning 
quality does not scale linearly with context length. Information positioned 
in the middle of the context is harder to retrieve~\cite{lost_in_middle}, 
irrelevant content actively dilutes the model's attention toward 
decision-critical evidence~\cite{lost_in_middle, multi_turn_lost}, 
and the effective context window is substantially smaller than its nominal 
size~\cite{anthropic_context}. These are not artifacts of any 
particular model, but intrinsic properties of current LLM architectures. 
The implication is direct. As long as an agent uses an LLM as its reasoning 
engine, the quality of each decision step is ultimately determined within a 
single forward pass, and no amount of tooling, memory capacity, or workflow 
complexity can circumvent this constraint. Context information density is 
therefore not an optional optimization target, but a structural constraint 
that every agent system must confront by design.

\textbf{There exists a minimal complete capability set for agent systems.} 
Under the structural constraint of information density, our experience 
suggests that an agent needs to implement only three capabilities, namely 
\textbf{tool interfacing}, \textbf{context management}, and \textbf{memory 
formation}. These three capabilities correspond to the three inevitable 
stages in the agent task execution pipeline where information density is 
systematically degraded, and therefore constitute what we propose as the 
minimal set of capabilities an agent framework must implement. 
(1) Tool interfacing is the sole channel through 
which an agent interacts with the external world. Redundant tool definitions 
consume large portions of the context budget before task execution begins, 
so interface design must be deliberately constrained to keep overhead 
minimal. (2) Context management corresponds to the input to the language 
model. Task state, intermediate results, tool outputs, and all other content 
must be actively filtered before entering the context, as loading everything 
by default allows irrelevant information to crowd out critical evidence. 
(3) Memory formation corresponds to cross-task knowledge accumulation. 
Without retaining verified content from interaction into reusable memory, 
every task begins from scratch. Any additional complexity that does not 
serve one of these three capabilities is, in our view, actively degrading 
information density.

\textbf{In agent systems, lower token consumption corresponds to better 
task performance.} This finding is counterintuitive, since the prevailing 
assumption is that longer reasoning chains and more interaction turns reflect 
more thorough deliberation, and should therefore yield better outcomes. 
However, our experimental results systematically point to the opposite 
conclusion within the setting of long-horizon agentic execution. 
On Lifelong AgentBench, GA consumes only 
27.7\% of Claude Code's input tokens and 15.5\% of OpenClaw's, while 
achieving a higher task completion rate of 100\%. This pattern holds 
consistently across multiple benchmarks. As discussed above, beyond a 
certain point, additional tokens do not introduce more useful information 
but instead degrade reasoning quality through positional bias, attention 
dilution, and effective window contraction~\cite{lost_in_middle, 
multi_turn_lost, anthropic_context}. An agent that consumes more tokens 
is more likely suffering from systematic failures in context management, 
compensating for degraded per-step decision quality through additional 
interactions rather than improving it. We therefore propose that, in 
long-horizon agentic settings, token consumption reflects the symptoms 
of an agent's context management quality, rather than the thoroughness 
of reasoning.

\textbf{Permissions define the ceiling of agent capability.} What an agent 
can perceive, what it can act upon, and what feedback it can learn from 
directly determine the complexity of reasoning chains it can develop and 
the difficulty of tasks it can solve. The scope of permissions is the 
boundary of the capability development space, which cannot be decoupled. 
Locking down the action boundary during the agent's exploration phase is 
equivalent to preemptively capping its capability ceiling at the system 
design stage. An agent restricted to reading a small set of files, unable 
to execute code or access external information, can only operate within a 
truncated state space regardless of how capable its underlying model is. 
Narrowing the exploration boundary is not a path toward building useful 
agents, the endpoint of which is a system that is safe, but useless.

\textbf{Minimal architecture is a necessary prerequisite for autonomous 
agent evolution.} In practice, we find that GA's autonomous exploration 
already consolidates far more skills than manual intervention could 
accumulate, representing a first step beyond traditional human-in-the-loop 
paradigms. We conjecture that the implications of architectural minimality 
extend further still. When the architecture is sufficiently minimal, the 
target of evolution can expand from skills to the architecture itself. A 
system with hundreds of thousands of lines of code is opaque to the 
agent---it can neither understand nor modify it. A core codebase of a few 
thousand lines, by contrast, is readable, understandable, and modifiable. 
In GA's minimal architecture, the self-hosted CLI as the native execution 
surface naturally enables subagents to read and modify the core codebase, 
making architectural self-update a practically achievable next step. Agent 
evolution thus presents three progressive dimensions: skill consolidation, 
autonomous exploration, and architectural self-update. We leave the 
validation of this full evolutionary path as an open problem for future 
work, but argue that minimal architecture is its necessary prerequisite.

%% file: section/5related.tex
5
Prior work on autonomous LLM agents has advanced several ingredients of long-horizon execution, including action interfaces, memory, self-improvement, and web interaction. Our position relative to this literature is more specific than a generic ``integrated system'' claim. GA is built around a single systems objective, \emph{maximizing contextual information density}, and studies how multiple agent components should be co-designed when the bottleneck is not raw capability but how much decision-relevant information can be preserved within a limited context budget. In this sense, GA is not simply a new memory module, a new reflection method, or a new browser agent; it is a general-purpose agent architecture that connects these ingredients through a common optimization target.

\subsection{LLM-Based Agent Systems and Action Interfaces}
\label{sec:related_llm_agents}

LLM agents have evolved from prompt-level reasoning loops to integrated systems that execute long action sequences in real environments. ReAct~\citep{react} and Reflexion~\citep{reflexion} established the basic loop of interleaving reasoning, acting, and feedback. AutoGPT~\citep{autogpt} popularized iterative goal decomposition, while MetaGPT~\citep{metagpt} pushed the field toward explicit multi-role coordination and workflow design. This line established the key intuition that agent performance depends not only on the base model, but also on how deliberation is scaffolded over multiple steps.

As agents moved into software engineering and computer-use settings, the action interface itself became a first-class design choice. CodeAct~\citep{codeact} unifies agent actions as executable code, which increases compositional flexibility and makes behaviors easier to test and reuse. Devin~\citep{devin}, SWE-agent~\citep{sweagent}, and OpenHands~\citep{openhands} further show that performance depends strongly on how the model is connected to the external environment, whether through integrated coding workflows, specialized Agent--Computer Interfaces, or open agent runtimes. Recent product systems such as Claude Code~\citep{claudecode}, Codex~\citep{codex}, Manus~\citep{manus}, and OpenClaw~\citep{openclaw_rl} reinforce the same trend in practice.

GA is closest to this systems line, but differs in what it optimizes. Most existing agent systems primarily expand reachable behaviors through richer workflows, more specialized tools, or stronger environment integration. By contrast, GA asks how broad computer-use capability can be preserved while keeping the action space and prompt overhead deliberately small. Its minimal atomic toolset is therefore not just an implementation choice, but part of the paper's central claim: cross-domain long-horizon performance can improve when tool abstraction is designed to increase contextual information density rather than to maximize interface richness.

\subsection{Memory and Context Management}
\label{sec:related_memory}

A second line of work studies how agents retain only behaviorally useful information as trajectories grow. MemGPT~\citep{memgpt} treats the context window as a limited working memory and external storage as archival memory, introducing a paging-style view of agent memory. A-MEM~\citep{amem} instead models memory as a dynamically evolving network of atomic notes and links, enabling richer associative recall. These approaches make long-horizon agents more feasible, but they focus primarily on storage and retrieval.

Recent work increasingly treats context construction itself as a systems problem. LongLLMLingua~\citep{longllmlingua} shows that prompt compression can preserve task-relevant information while reducing long-context cost, and practical context-engineering analyses emphasize that agent quality depends not only on window length but also on what enters the window and in what form~\citep{anthropic_context}. Production systems such as Claude Code~\citep{claudecode} and Manus~\citep{manus} similarly rely on artifact-based tracking and periodic compaction to extend effective horizon.

GA differs from both memory-centric and compression-centric approaches in two ways. First, it treats memory quality as a \emph{verification and selection} problem, not only a storage or retrieval problem: only behavior-changing and validated information is promoted into longer-term representations. Second, it optimizes the full path from observation to retained memory, rather than only the final retrieval step. The resulting design goal is not to keep more history accessible, but to keep the active context as sparse, reliable, and decision-relevant as possible.

\subsection{Self-Evolution and Experience Distillation}
\label{sec:related_self_evolution}

Research on self-evolving agents asks how repeated execution can become future capability. A recent survey~\citep{selfevolvesurvey} organizes the space by what evolves, when evolution happens, and what feedback drives improvement. Within this space, Agent-Pro~\citep{agentpro} studies policy-level reflection and optimization, showing that agents can revise their own behavioral policies without updating model parameters. Voyager~\citep{voyager} demonstrates a stronger form of accumulation in a specialized environment by continually storing verified executable skills.

Broader general-purpose work mostly evolves the agent through textual abstractions. EvolveR~\citep{evolver}, FLEX~\citep{flex}, AgentEvolver~\citep{agentevolver}, and experience-driven lifelong learning~\citep{selfevolvinglifelong} all convert trajectories into strategic principles, reflections, or structured knowledge units that help later execution. This is an important step beyond simple history reuse, but in most cases the retained experience remains natural-language guidance rather than executable capability.

GA is aligned with this line in viewing reflection as the engine of continual improvement, but it makes a stronger systems claim: the endpoint of reflection should be a \emph{representation shift} from verbose trajectories to compact operational assets. Verified experience is progressively transformed into SOPs, code, and reusable skills, so improvement appears not only as better policy guidance but also as lower inference-time cost. This is why our evaluation emphasizes efficiency convergence and token reduction under repeated execution, rather than only one-shot task success.

%% file: section/7conclusion.tex

We present \textbf{GenericAgent (GA)}, a self-evolving general-purpose LLM agent built around a single design principle: \textbf{context information density maximization}. Instead of treating context as a passive byproduct of interaction, GA optimizes both completeness and conciseness. It achieves this through four components: a minimal atomic tool set, a hierarchical on-demand memory, a reflection-driven self-evolution pipeline that distills verified trajectories into reusable SOPs and executable code, and a context truncation and compression layer that preserves information density during long execution.

Across five evaluation dimensions, GA shows a strong efficiency–performance trade-off. It matches or outperforms existing agent systems on task completion while using fewer tokens and interactions. Its self-evolution process also reduces token usage by up to 89.6\% across repeated runs, while maintaining or improving performance. Overall, these results suggest that improving long-horizon agent capability is not mainly about adding more tools, memory, or longer context. It is more about controlling how information is represented and maintained during execution. We release GA as an open-source system and hope this work can inform future research on general-purpose self-evolving agents.

%% file: section/appendix.tex
\definecolor{caseblueframe}{RGB}{67,119,183}
\definecolor{casebluebg}{RGB}{231,239,250}
\definecolor{casegrayframe}{RGB}{120,120,120}
\definecolor{casegraybg}{RGB}{243,243,243}

\newtcolorbox{CasePromptBox}[1]{
  breakable,
  colback=casebluebg,
  colframe=caseblueframe,
  colbacktitle=caseblueframe,
  coltitle=white,
  fonttitle=\bfseries,
  title={#1}
}

\newtcolorbox{CaseTraceBox}[1]{
  breakable,
  colback=casebluebg,
  colframe=caseblueframe,
  colbacktitle=caseblueframe,
  coltitle=white,
  fonttitle=\bfseries,
  title={#1}
}

\newtcolorbox{CaseAnswerBox}[1]{
  breakable,
  colback=casegraybg,
  colframe=casegrayframe,
  colbacktitle=casegrayframe,
  coltitle=white,
  fonttitle=\bfseries,
  title={#1}
}

\setcounter{subsection}{0}
\setcounter{subsubsection}{0}
\renewcommand{\thesubsection}{\arabic{subsection}}
\renewcommand{\thesubsubsection}{\thesubsection.\arabic{subsubsection}}

\subsection{Atomic Tool Alignment}
\label{app:atomic_tool_alignment}

This subsection provides a detailed capability-level alignment for the core atomic tools discussed in the main text.
The purpose is not to enumerate the full runtime tool inventories of Claude Code or OpenClaw, but to show that the core capabilities retained by GA each have corresponding prototypes in both systems.

\begin{table*}[htbp]
\centering
\caption{\textbf{Capability-level alignment of the basic tooling environment.} Each row maps one capability category to the corresponding tool in Claude Code, OpenClaw, and GA.}
\def\arraystretch{0.99}
\setlength{\tabcolsep}{0.42em}
\resizebox{1.0\linewidth}{!}{
\begin{tabular}{llll}
\toprule
\textbf{Capability Category} & \multicolumn{3}{c}{\textbf{Corresponding Tool / Mechanism}} \\
\cmidrule(lr){2-4}
 & \textbf{Claude Code} & \textbf{OpenClaw} & \textbf{GA} \\
\midrule
Code execution & BashTool & exec / process / nodes & code\_run \\
File reading & FileReadTool & read & file\_read \\
File writing & FileWriteTool & write & file\_write \\
Local file editing & FileEditTool & write / patch-like editing & file\_patch \\
Web reading  & WebFetchTool / browser read & browser base reading & web\_scan \\
Web interaction & WebBrowserTool / browser actions & browser interaction & web\_execute\_js \\
Working memory & TodoWriteTool / BriefTool & session history / status-like ability & update\_working\_checkpoint \\
\bottomrule
\end{tabular}
}
\label{tab:appendix_tool_alignment}
\end{table*}

Table~\ref{tab:appendix_tool_replacement_examples} complements the capability-level mapping above with representative task-level replacement examples.
Rather than claiming that GA reproduces every specialized tool one by one, the point is that many benchmarked capabilities can be reconstructed through short compositions of a few atomic tools.

\begin{table*}[t]
\centering
\small
\def\arraystretch{1.02}
\setlength{\tabcolsep}{0.35em}
\begin{tabularx}{\textwidth}{@{}>{\raggedright\arraybackslash}p{0.12\textwidth}>{\raggedright\arraybackslash}p{0.18\textwidth}>{\raggedright\arraybackslash}X>{\raggedright\arraybackslash}X>{\raggedright\arraybackslash}X@{}}
\toprule
\textbf{Source} & \textbf{Specialized Tool / Capability} & \textbf{Representative Task} & \textbf{GA Replacement Composition} & \textbf{Argument} \\
\midrule
Claude Code & \texttt{GlobTool} & \texttt{task\_cc\_01\_glob\_markdown\_list} & \texttt{code\_run} & File-pattern matching can be completed through general code execution. \\
Claude Code & \texttt{GrepTool} & \texttt{task\_cc\_02\_grep\_todo\_hits} & \texttt{code\_run} + \texttt{file\_read} & Text search and localization can be implemented by combining general execution with file reading. \\
\cmidrule(lr){1-5}
Claude Code & \texttt{WebFetchTool} & \texttt{task\_cc\_03\_webfetch\_example\_brief} & \texttt{web\_scan} + \texttt{web\_execute\_js} + \texttt{code\_run} & Web fetching, cleaning, and summarization can be reconstructed by combining browser primitives with code-based post-processing. \\
Claude Code & \texttt{AgentTool} & \texttt{task\_cc\_04\_agent\_fact\_extract} & \texttt{file\_read} + \texttt{code\_run} + \texttt{update\_working\_checkpoint} & Explicit agent delegation can be approximated through stored subagent strategies in memory together with general execution triggers. \\
\cmidrule(lr){1-5}
Claude Code & \texttt{WebSearchTool} & \texttt{task\_cc\_05\_websearch\_example\_results} & \texttt{web\_execute\_js} + \texttt{web\_scan} & Online search can be completed through browser interaction plus webpage reading. \\
Claude Code & \texttt{NotebookEditTool} & \texttt{task\_cc\_06\_notebook\_inspect} & \texttt{file\_read} + \texttt{code\_run} & Notebook-structure inspection can be handled by reading the \texttt{.ipynb} JSON and executing a parsing script. \\
\cmidrule(lr){1-5}
OpenClaw & \texttt{nodes} / structured data & \texttt{task\_oc\_02\_csv\_stats} & \texttt{code\_run} & CSV statistics can be completed through general code execution without a dedicated structured-data tool. \\
\bottomrule
\end{tabularx}
\caption{\textbf{Representative task-level substitutions from specialized tools to GA atomic-tool compositions.} Each row shows one concrete capability test, the original specialized tool or mechanism used by the baseline system, and the corresponding GA composition that can complete the same class of task.}
\label{tab:appendix_tool_replacement_examples}
\end{table*}

\FloatBarrier

\subsection{Web Browsing Visualization}
\label{app:web_browsing_visualization}

\begin{figure}[htbp]
\centering
\includegraphics[width=0.7\linewidth]{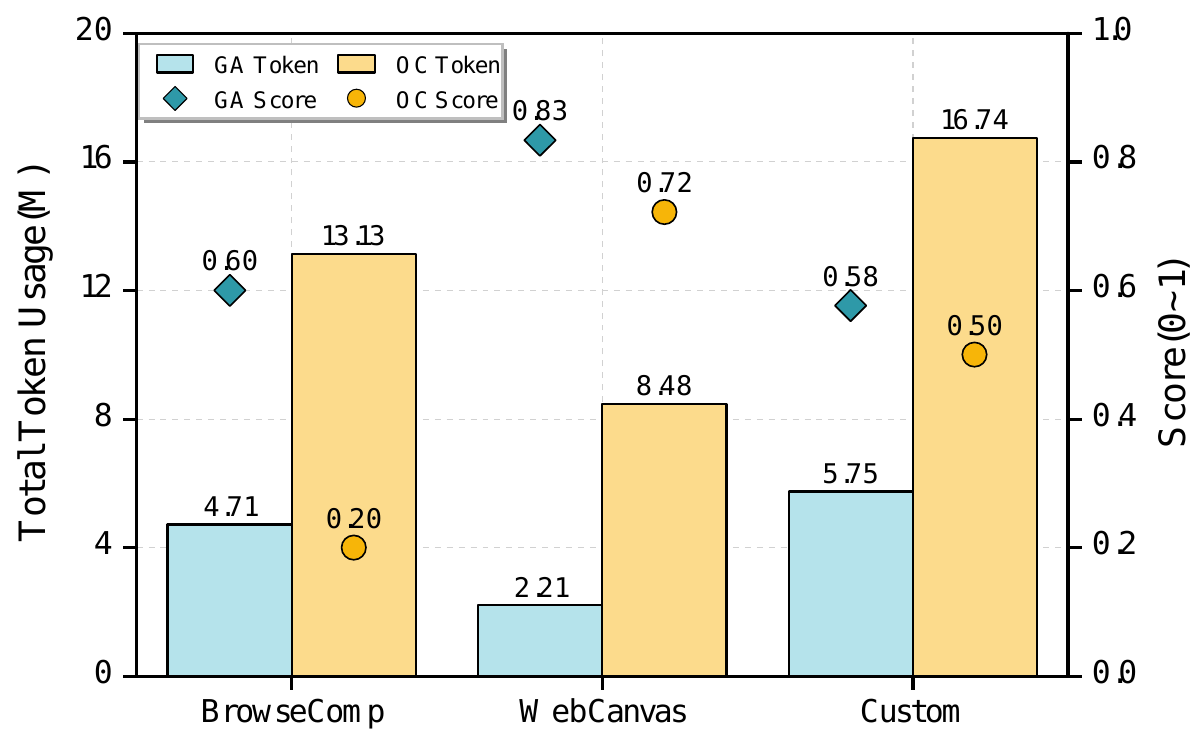}
\caption{\textbf{Token consumption and normalized score across three web browsing benchmarks.} The left axis shows total token consumption in millions (M), while the right axis shows normalized scores on a 0--1 scale. GA consistently achieves competitive or superior performance while consuming significantly fewer tokens than OpenClaw across all three benchmarks.}
\label{fig:web_browsing_dual_axis}
\end{figure}

Figure~\ref{fig:web_browsing_dual_axis} provides a visual comparison of token consumption and normalized performance across the three web browsing benchmarks discussed in Section~\ref{sec:exp_browsing}.

\FloatBarrier

\subsection{Case Studies}
\label{app:case_studies}

We present four representative cases corresponding to the main evaluation dimensions.
For readability, the cases are ordered as tool use, memory, self-evolution, and web browsing.
Each case follows a ``purpose and task setup $\rightarrow$ artifacts / trace $\rightarrow$ observed outcome'' pattern so that the appendix reads as a set of worked examples rather than a list of compressed summaries.

\begin{itemize}[leftmargin=*]
\item \textbf{Case A1 (Tool Use).} An API procurement workflow requiring web lookup, structured extraction, budget reasoning, and file generation, used to show how GA handles a long-horizon task with a minimal atomic-tool set (Section~\ref{sec:dimension4_tool_use_efficiency}).

\item \textbf{Case A2 (Memory).} Dangerous-goods hazard classification under condensed, full, and redundant memory variants, used to show why information density matters more than information volume (Section~\ref{sec:exp_memory}).

\item \textbf{Case A3 (Self-Evolution).} A GitHub PR research task showing the progression from natural-language SOP to compiled Python code and a transferability note, used to demonstrate how GA converts experience into reusable executable assets (Section~\ref{sec:self_evolution_capability}).

\item \textbf{Case A4 (Web Browsing).} A BrowseComp-ZH multi-hop reasoning chain where GA identifies a historical figure through iterative search refinement, used to illustrate structured browser extraction in practice (Section~\ref{sec:exp_browsing}).
\end{itemize}

\subsubsection{Case A1. Tool Use: API Procurement Workflow}
\label{app:case_tools}

\begin{CasePromptBox}{Purpose and Task Setup}
\small\ttfamily
Purpose: to examine whether GA's atomic-tool design can support a long-horizon, multi-step workflow, \\
we provide a laboratory procurement task over public LLM pricing pages. \\
This task is useful because it requires browsing, structured extraction, numerical reasoning, and file generation in one workflow. \\
The goal of the case is to show that GA's minimal atomic-tool set can complete the full chain and reach the correct recommendation, providing a qualitative counterpart to Table~\ref{tab:appendix_tool_case}. \\
Task: investigate laboratory procurement for text-only LLM APIs using these public pricing pages: \\
- OpenAI API pricing: https://openai.com/api/pricing/ \\
- Anthropic pricing: https://www.anthropic.com/pricing \\
- Gemini API pricing: https://ai.google.dev/gemini-api/docs/pricing \\
Constraints: \\
- only text models are considered \\
- laboratory size: 6 students \\
- monthly input tokens: 80M \\
- monthly output tokens: 20M \\
- budget: \$300 / month \\
Requirements: \\
1. record candidate flagship text models and their input/output prices \\
2. test whether a single-model plan fits the budget \\
3. if not, trigger fallback and test a dual-model plan \\
4. if that still fails, output infeasible \\
5. final answer must state the primary option, whether it is feasible, whether fallback was triggered, the recommended plan, the estimated monthly cost, and the reason \\
Required output files: cost\_comparison.csv and decision\_04.json.
\end{CasePromptBox}

\begin{CaseTraceBox}{Artifact 1: Cost Comparison File}
\small\ttfamily
cost\_comparison.csv \\
provider,model,input\_cost\_per\_1m,output\_cost\_per\_1m,estimated\_monthly\_cost\_usd,meets\_budget \\
OpenAI, GPT-5.4, 2.5, 15.0, 500.0, No \\
OpenAI, GPT-5.4 mini, 0.75, 4.5, 150.0, Yes \\
OpenAI, GPT-5.4 nano, 0.2, 1.25, 41.0, Yes \\
Anthropic, Claude Opus 4.6, 5.0, 25.0, 900.0, No \\
Anthropic, Claude Sonnet 4.6, 3.0, 15.0, 540.0, No \\
Anthropic, Claude Haiku 4.5, 1.0, 5.0, 180.0, Yes \\
Google, Gemini 2.5 Pro, 1.25, 10.0, 300.0, Yes \\
Google, Gemini 2.5 Flash, 0.3, 2.5, 74.0, Yes \\
Google, Gemini 2.5 Flash-Lite, 0.1, 0.4, 16.0, Yes
\end{CaseTraceBox}

\begin{CaseTraceBox}{Artifact 2: Decision File}
\small\ttfamily
primary\_option: Gemini 2.5 Pro \\
primary\_option\_feasible: true \\
fallback\_triggered: false \\
recommended\_plan: single\_model \\
recommended\_models: [Gemini 2.5 Pro] \\
estimated\_monthly\_cost\_usd: 300.0 \\
reason: Gemini 2.5 Pro is a strong single-model option that still fits within the budget. \\
Cost calculation: 80M input x \$1.25/MTok + 20M output x \$10.00/MTok = \$300.0/month. \\
Other flagship models exceed the budget: GPT-5.4 at \$500/month and Claude Opus 4.6 at \$900/month. \\
No fallback to a dual-model plan is needed.
\end{CaseTraceBox}

\begin{CaseAnswerBox}{Observed Outcome}
\small\ttfamily
GA completed the task in 21 requests and 364,385 total tokens (324.6s). \\
All grading criteria passed: cost\_comparison.csv and decision\_04.json both created with correct schema. \\
Final recommendation: Gemini 2.5 Pro as a single-model plan at exactly \$300/month. \\
Cost breakdown: 80M input $\times$ \$1.25/MTok + 20M output $\times$ \$10.00/MTok = \$300.0. \\
Other flagship models exceeded the budget: GPT-5.4 at \$500/month, Claude Opus 4.6 at \$900/month. \\
No fallback to a dual-model plan was needed.
\end{CaseAnswerBox}

\begin{table}[t]
\centering
\caption{Cross-system comparison on the long-horizon procurement task.}
\label{tab:appendix_tool_case}
\begin{tabular}{lcccc}
\toprule
\textbf{System} & \textbf{Requests} & \textbf{Total Tokens} & \textbf{Time (s)} & \textbf{Success} \\
\midrule
GenericAgent & 21 & 364,385 & 324.64 & Yes \\
Claude Code & 29 & 485,335 & 224.53 & Yes \\
OpenClaw & 12 & 428,402 & 108.42 & Yes \\
\bottomrule
\end{tabular}
\end{table}

\FloatBarrier

\subsubsection{Case A2. Memory: Dangerous-Goods Hazard Classification}
\label{app:case_memory}

\begin{CasePromptBox}{Purpose and Task Setup}
\small\ttfamily
Purpose: to examine whether memory quality depends more on information density than on raw information volume, \\
we provide a dangerous-goods classification task under three memory variants. \\
This task is useful because the decision rule stays fixed while only the memory representation changes. \\
The goal of the case is to show that the main effect comes from how the rules are packaged and retrieved, rather than from whether the rules exist at all. \\
Task: determine hazard\_class from product\_id and the four source texts. \\
Inputs: \\
- SDS label text \\
- handling/storage guidance \\
- transportation requirements \\
- disposal guidance \\
The output must contain both hazard\_score and hazard\_class.
\end{CasePromptBox}

\begin{CaseTraceBox}{Artifact 1: Condensed Memory}
\small\ttfamily
Required rules: \\
1. First validate product\_id. It must be resolvable and follow format P\_XXXXX. \\
\quad If invalid, stop and output hazard\_score=0 and hazard\_class="Unable to Decide". \\
2. Get four component scores: \\
\quad - safety score from SDS label text \\
\quad - handling score from handling/storage guidelines \\
\quad - transportation score from transportation requirements \\
\quad - disposal score from disposal guidelines \\
\quad Each valid component score must be in [1,5]. \\
3. Compute hazard\_score: \\
\quad - If one component is missing or 0, replace it with the max of the other available scores. \\
\quad - If more than two components are missing or 0, output hazard\_score=0 and hazard\_class="Unable to Decide". \\
\quad - Otherwise hazard\_score = sum of the four component scores. \\
\quad - Final valid range is 4 to 20. \\
4. Map hazard\_score to class: \\
\quad - 4-7 to Hazard Class A \\
\quad - 8-12 to Hazard Class B \\
\quad - 13-16 to Hazard Class C \\
\quad - 17-20 to Hazard Class D \\
5. Final output must include both hazard\_score and hazard\_class.
\end{CaseTraceBox}

\begin{CaseTraceBox}{Artifact 2: Full Memory}
\small\ttfamily
1. Purpose \\
To establish a standardized methodology for the systematic identification and classification of dangerous goods hazard classes through multi-source data integration and quantitative severity assessment protocols. \\
2. Scope \\
This procedure encompasses all dangerous goods shipment classification processes within the organization's supply chain operations. \\
3. Definitions \\
SDS = Safety Data Sheet \\
HS = Handling and Storage Guidelines \\
TR = Transportation Requirements \\
DG = Disposal Guidelines \\
AIP = API Integration Protocol \\
HCM = Hazard Classification Matrix \\
SAS = Severity Assessment Score \\
4. Input \\
Product ID (format: P\_XXXXX) \\
Source documentation: SDS / Handling and Storage / Transportation / Disposal \\
API access credentials: endpoint URLs / authentication tokens / backup authentication protocols \\
5. Main Procedure \\
Validate product identification documentation completeness. \\
If product ID fails format requirements, no further action should be taken. \\
The hazard score should be marked as 0 and hazard class as Unable to Decide. \\
Extract four scores, each between 1 and 5. \\
hazard\_score = safety score + handling score + transportation score + disposal score \\
If any component is missing or 0, impute it by taking the max of the other scores. \\
If more than two component scores are missing, mark Unable to Decide. \\
6. Output \\
Final hazard class designation \\
Digital record in Hazard Classification Registry \\
API response logs for all scoring calculations \\
Classification audit trail documentation \\
Final output should be in XML format with tags \textless{}hazard\_score\textgreater{} and \textless{}hazard\_class\textgreater{}
\end{CaseTraceBox}

\begin{CaseTraceBox}{Artifact 3: Redundant Memory}
\small\ttfamily
You are following a standardized dangerous-goods classification procedure designed for supply-chain compliance and multi-source hazard assessment. \\
Definitions: \\
- SDS = safety data sheet \\
- HS = handling and storage guidance \\
- TR = transportation requirements \\
- DG = disposal guidelines \\
- SAS = severity assessment score from 1 to 5 \\
- HCM = hazard classification matrix \\
Input scope: product\_id, SDS text, handling/storage text, transportation text, disposal text, API access and validation context. \\
Decision workflow: \\
1. Validate documentation completeness and product identifier validity. \\
2. product\_id must match format P\_XXXXX and be resolvable. \\
3. If invalid, do not continue. Set hazard\_score to 0 and hazard\_class to ``Unable to Decide''. \\
4. Analyze SDS, handling/storage, transportation, and disposal sources separately. \\
5. Derive one component score per source, each between 1 and 5. \\
6. If exactly one component is missing or 0, impute it using the maximum of the other component scores. \\
7. If more than two components are missing, return ``Unable to Decide'' and hazard\_score 0. \\
8. Sum the four components into a cumulative hazard\_score and validate that total score lies in the range 4-20. \\
9. Convert the total score into Hazard Class A / B / C / D. \\
10. Return the final result in structured form with hazard\_score and hazard\_class, preserving an audit-ready trail. \\
Background note: the broader SOP also mentions registry logging, API integration, source-document handling, and record retention, but the operationally decisive rules are the identifier check, missing-value handling, score summation, and class mapping.
\end{CaseTraceBox}

\begin{CaseAnswerBox}{Observed Outcome}
The memory ablation becomes visually intuitive once the three variants are shown side by side.
The condensed file places every behavior-changing rule in a single short block, whereas the full file spreads the same operational core across purpose statements, definitions, credential notes, registry requirements, and output-format details.
Table~\ref{tab:memory_ablation} can therefore be read primarily as an information-placement result rather than a simple information-volume result.
\end{CaseAnswerBox}

\FloatBarrier

\subsubsection{Case A3. Self-Evolution: GitHub PR Research}
\label{app:case_self_evolution}

\begin{CasePromptBox}{Purpose and Task Setup}
\small\ttfamily
Purpose: to examine whether repeated execution can be distilled into reusable assets, \\
we provide a GitHub PR research task that evolves from an SOP into executable code. \\
This task is useful because the representation shift is directly observable: the agent first stabilizes a procedure as text and then compiles that procedure into code. \\
The goal of the case is to show that what changes across iterations is not merely token count, but the form of reusable knowledge, which serves as the qualitative counterpart of Table~\ref{tab:self_evolution_rounds}. \\
Investigate the five most recent merged bug-fix PRs in a GitHub repository. \\
For each PR: recover the linked issue, identify the affected module, and verify whether the module has troubleshooting coverage in the documentation site. \\
Produce a structured JSON report.
\end{CasePromptBox}

\begin{CaseTraceBox}{Artifact 1: SOP}
\small\ttfamily
Scenario: batch investigation of GitHub PRs, including PR details, linked issues, affected modules, and documentation coverage. \\
Core strategy: \\
DO NOT browse PR pages one by one -- efficiency is extremely low and the run may exceed the turn budget. \\
PREFER batch extraction -- use a script or API to obtain multiple PR records at once. \\
Recommended solution: use github\_pr\_analyzer.py to complete the whole chain in one pass. \\
Example command: \\
python3 ../memory/github\_pr\_analyzer.py langchain-ai/langchain \\
\quad --doc-url https://python.langchain.com \\
\quad --limit 5 \\
\quad --output report.json \\
Script advantages: \\
- pure Python, no browser environment required \\
- one-click execution of the full workflow \\
- flexible command-line parameters \\
- built-in error handling \\
Manual fallback flow: \\
1. build a filtered PR URL \\
2. use document.querySelectorAll(.js-issue-row) to extract PR numbers and titles \\
3. batch fetch PR HTML \\
4. recover issue links with the rule /issues/\textbackslash d+, then Fixes/Closes \#\textbackslash d+, then standalone \#\textbackslash d+ \\
5. infer module names from PR title prefixes such as community: \\
6. verify documentation coverage against troubleshooting and integration paths \\
7. dump the final report with json.dump() \\
Pitfalls: \\
1. do not use window.location.href to visit PR pages one by one \\
2. prefer browser fetch() when manual HTML collection is needed \\
3. distinguish issue links from pull-request links \\
4. different projects may use different documentation subdomains
\end{CaseTraceBox}

\begin{CaseTraceBox}{Artifact 2: Compiled Code}
\small\ttfamily
\textbf{fetch\_pr\_list(...)} \\
\quad url = https://github.com/\{repo\}/pulls?q=\{filters\} \\
\quad response = session.get(url, timeout=30) \\
\quad soup = BeautifulSoup(response.text, 'html.parser') \\
\quad pr\_elements = soup.select('.js-issue-row') \\
\quad collect number, title, and url for the first limit PRs \\
\textbf{extract\_pr\_details(pr)} \\
\quad \detokenize{module_match = re.match(r'^([^:\[]+)', pr['title'])} \\
\quad pattern 1: full GitHub issue URL \\
\quad pattern 2: Fixes|Closes|Resolves \#N \\
\quad pattern 3: standalone \#N excluding the PR number itself \\
\quad return pr\_number, bug\_module, issue\_link \\
\textbf{check\_doc\_coverage(module)} \\
\quad probe multiple candidate paths: \\
\quad /docs/troubleshooting/ \\
\quad /docs/troubleshooting/\{module\} \\
\quad /docs/integrations/\{module\}/ \\
\quad /docs/modules/\{module\}/ \\
\quad /api\_reference/\{module\}/ \\
\quad /docs/integrations/providers/\{module\}/ \\
\quad accept the module if the page contains both the module name and troubleshooting \\
\textbf{analyze(...)} \\
\quad fetch PR list, then extract details, then check documentation, then call json.dump(results, f, indent=2, ensure\_ascii=False)
\end{CaseTraceBox}

\begin{CaseTraceBox}{Artifact 3: Transferability Note}
\small\ttfamily
Original task core abilities: \\
- batch retrieval of GitHub PR lists with filters \\
- extraction of PR details, titles, modules, and linked issues \\
- verification of external documentation coverage \\
- generation of structured JSON reports \\
Reusable asset set: \\
- SOP document: github\_pr\_research\_sop.md \\
- Python script: github\_pr\_analyzer.py \\
Transfer targets described in the case note: \\
1. GitHub issue batch research \\
2. contributor analysis \\
3. release tracking \\
4. workflow and CI inspection \\
5. code search and pattern extraction \\
Example reuse in the issue-analysis case: \\
- keep the batch-fetch strategy \\
- replace /repos/\{repo\}/pulls with /repos/\{repo\}/issues \\
- reuse regex extraction and concurrent request handling \\
Estimated adaptation cost in the transferability note: about 30\% code changes, mainly endpoint replacement and field mapping.
\end{CaseTraceBox}

\begin{CaseAnswerBox}{Observed Outcome}
\small\ttfamily
LangChain verification case (2026-04-08): \\
python3 github\_pr\_analyzer.py langchain-ai/langchain --doc-url https://python.langchain.com --limit 5 \\
\\
5 merged bug-fix PRs retrieved \\
modules found: community (2), partners, openai, core/anthropic \\
issue links recovered: 4/5 \\
troubleshooting coverage found: 4/5 \\
runtime: about 30--40 seconds
\end{CaseAnswerBox}

\FloatBarrier

\subsubsection{Case A4. Web Browsing: BrowseComp-ZH Reasoning Chain}
\label{app:case_browsing}

\begin{CasePromptBox}{Purpose and Task Setup}
\small
Purpose: to examine multi-hop web browsing under a compressed browser interface, \\
we provide a BrowseComp-ZH question whose answer must be recovered through iterative query refinement. \\
This example is useful because the answer is not available from a single lookup and the chain of evidence must be assembled step by step. \\
The goal of the case is to make the narrowing process visible: identify the \emph{shihua} pioneer, confirm Ouyang Xiu, and then recover Wang Anshi from the quotation. \\
Original question (BrowseComp-ZH): \\
``There is a writer A who participated in a literary reform movement, \\
advocated a plain but disciplined style, opposed overly strange and risky writing, \\
and pioneered the \emph{shihua} form of poetic criticism. \\
Another contemporary writer B commented that he had ``deep character'' and ``broad knowledge''. \\
Who was writer B?'' \\
Recorded case metadata: dataset = browsecomp, score = 1.0, time = 130.4s, tokens = input 30050 / output 1296, turns = 7, tools = 6.
\end{CasePromptBox}

\begin{CaseTraceBox}{Artifact 1: Interaction Trace}
\small\ttfamily
Turn 1 summary: start by searching ``first shihua writer'' to identify writer A. \\
Action query: shihua inventor plain style \\
Turn 2 summary: scan the Google result page and narrow the candidate set. \\
Turn 3 summary: because the first search does not uniquely resolve the target, search ``Ouyang Xiu + Liuyi Shihua''. \\
Action query: Ouyang Xiu Liuyi Shihua \\
Turn 4 summary: confirm writer A as Ouyang Xiu. \\
Turn 5 summary: search the evaluative phrase together with Ouyang Xiu. \\
Action query: deep character broad knowledge Ouyang Xiu \\
Turn 6 summary: verify that the quoted evaluator is Wang Anshi. \\
Turn 7 summary: return the final answer with confidence.
\end{CaseTraceBox}

\begin{CaseTraceBox}{Artifact 2: Final Reasoning Excerpt}
\small\ttfamily
Explanation: \\
Through the searches on ``first shihua writer'' and related keywords, the agent identified writer A as Ouyang Xiu. \\
He participated in the Northern Song literary reform movement, advocated a plain but disciplined style, \\
opposed strange and risky writing, and pioneered the form through Liuyi Shihua. \\
After searching the evaluative phrase together with Ouyang Xiu, \\
the agent found that Wang Anshi, in a commemorative text for Ouyang Xiu, \\
described him as having ``deep character'' and ``broad knowledge''. \\
Therefore the writer B in the question is Wang Anshi.
\end{CaseTraceBox}

\begin{CaseAnswerBox}{Observed Outcome}
\small\ttfamily
score: 1.0 \\
turns: 7 \\
tool calls: 6 \\
time: 130.4s \\
Exact Answer: Wang Anshi \\
Judge result: correct = yes, confidence = 95\% \\
Reasoning note: the extracted final answer exactly matches the gold answer ``Wang Anshi''.
\end{CaseAnswerBox}

\FloatBarrier

\subsection{General Capability Showcase}
\label{app:general_capability_showcase}

Beyond the four evaluation-aligned cases above, we present five additional cases drawn from a pool of more than 500 historical sessions.
They are selected to maximize coverage of non-benchmark capability dimensions rather than to reflect average-case performance, and are intended as qualitative demonstrations rather than statistical evaluation.
All cases come from user-authorized sessions; sensitive identities, document contents, account details, and message contents are anonymized or paraphrased when presented here, and any autonomous behaviors were executed under user-configured triggers or preset operating policies.
Together, these cases illustrate how GA's architectural choices --- atomic tools, persistent memory, and self-evolution --- support real-world workflows that are difficult to capture in standardized benchmarks.
Each case follows the same ``purpose and task setup $\rightarrow$ execution trace $\rightarrow$ observed outcome'' format for consistency.

\begin{itemize}[leftmargin=*]
\item \textbf{Case B1 (Cross-Device Control).} A mobile food-ordering task via ADB, followed by screen-recording retrieval and video post-processing, used to show how atomic-tool composition can bridge the PC--smartphone boundary within one execution chain.

\item \textbf{Case B2 (Cross-Platform Orchestration).} A message-forwarding task from a local WeChat database to a Weibo post, used to show how GA composes heterogeneous local and web environments within a single workflow.

\item \textbf{Case B3 (Autonomous Operation).} An overnight session triggered after the user leaves, used to show how self-evolution, working checkpoints, and preset patrol routines can support bounded autonomous operation.

\item \textbf{Case B4 (Remote Infrastructure).} An SSH-based remote file-server deployment task, used to show how atomic tools can support end-to-end DevOps workflows including dependency installation, file transfer, service deployment, and iterative troubleshooting.

\item \textbf{Case B5 (Long-Horizon Academic Workflow).} A multi-session NSFC grant proposal assistance task spanning figure design, citation verification, and batch error correction, used to show how persistent memory supports extended, multi-phase academic work.
\end{itemize}

\subsubsection{Case B1. Cross-Device Control: Mobile Food Ordering via ADB}
\label{app:case_cross_device}

\begin{CasePromptBox}{Purpose and Task Setup}
\small\ttfamily
Purpose: to examine whether GA can operate across the PC--smartphone boundary through ADB-based device control, \\
we provide a real-world mobile food-ordering task followed by screen-recording retrieval and video post-processing. \\
This task is useful because it requires coordinated control of a physical mobile device, GUI interaction with a commercial app, and multimedia processing on the host PC within a single workflow. \\
The goal of the case is to show that a small atomic-tool set can be composed into a cross-device execution chain under a realistic consumer-app workflow. \\[6pt]
User instruction (verbatim, translated): \\
``Order two cups of milk tea on my phone via Meituan Waimai. Stop at the final payment page --- do not pay.'' \\[4pt]
Follow-up instructions: \\
1. ``Pull the screen recording from the phone.'' \\
2. ``Trim everything before 1:10, black out the top half after 3:40.'' \\
3. ``Convert it into a sped-up GIF.''
\end{CasePromptBox}

\begin{CaseTraceBox}{Artifact 1: Mobile Ordering Execution Trace}
\small\ttfamily
Step 1: Establish ADB connection to the Android device. \\
Step 2: Launch the Meituan Waimai (food delivery) app via ADB shell am start. \\
Step 3: Dismiss pop-up advertisements by locating and tapping the close button coordinates. \\
Step 4: Navigate to the ``Desserts \& Drinks'' category. \\
Step 5: Select a nearby store (Hushang Ayi, a popular milk-tea chain). \\
Step 6: Locate the ``Must-Try Milk Tea --- Pick Any Two'' combo at \textyen 23.9. \\
Step 7: Select flavor 1: Thick Taro Paste; select flavor 2: Brown Sugar Boba. \\
Step 8: Confirm selections and add to cart. \\
Step 9: Proceed to the checkout page and halt --- no payment action taken.
\end{CaseTraceBox}

\begin{CaseTraceBox}{Artifact 2: Video Post-Processing Trace}
\small\ttfamily
Step 1: Pull the screen recording from the phone via adb pull. \\
Step 2: Trim the first 1 minute 10 seconds using ffmpeg -ss 00:01:10. \\
Step 3: Black out the top half of the frame after the 3:40 mark using ffmpeg drawbox filter with overlay logic. \\
Step 4: Re-encode and speed up the video (4x) to produce a compact GIF using ffmpeg with palettegen and paletteuse filters. \\
Step 5: Verify the final GIF file size and frame count.
\end{CaseTraceBox}

\begin{CaseAnswerBox}{Observed Outcome}
\small\ttfamily
In this session, GA completed the full pipeline: ADB device connection, commercial app navigation (7 GUI interactions), screen-recording retrieval, video trimming, region-specific masking, and GIF generation. \\
The workflow was executed through a composition of code\_run (ADB commands, ffmpeg), file\_read / file\_write (media files), and update\_working\_checkpoint (tracking multi-phase progress). \\
No task-specific mobile-automation framework (e.g., Appium, UIAutomator) was introduced in this workflow. \\
This case specifically illustrates the paper's claim that a compact atomic-tool set can be composed into a long execution chain that extends beyond the host PC when ADB is available as a bridge.
\end{CaseAnswerBox}

\FloatBarrier

\subsubsection{Case B2. Cross-Platform Orchestration: WeChat to Weibo Message Forwarding}
\label{app:case_cross_platform}

\begin{CasePromptBox}{Purpose and Task Setup}
\small\ttfamily
Purpose: to examine whether GA can orchestrate data flow across heterogeneous local and web platforms, \\
we provide a task that requires reading from a local encrypted database and publishing to a social-media platform via browser automation. \\
This task is useful because it chains together local database decryption, contact resolution, message extraction, and browser-based content publishing --- capabilities that belong to different software stacks. \\
The goal of the case is to show that GA can compose heterogeneous data sources and output channels into a single workflow under explicit user instruction. \\[6pt]
User instruction (verbatim, translated): \\
``Forward the latest message from Professor Xiao in WeChat to Weibo.'' \\[4pt]
Presentation note: contact identity and forwarded message content are anonymized here, and the operation was performed only after the user explicitly requested the forwarding action.
\end{CasePromptBox}

\begin{CaseTraceBox}{Artifact 1: Execution Trace}
\small\ttfamily
Phase 1 --- WeChat Message Extraction: \\
Step 1: Locate the local WeChat database files (EnMicroMsg.db). \\
Step 2: Decrypt the database using SQLCipher with the derived key. \\
Step 3: Query the contact table to resolve the anonymized alias ``Professor Xiao'' to the intended contact record. \\
Step 4: Query the message table for the most recent private message from this contact. \\
Step 5: Extract the latest message content (paraphrased here for privacy). \\[4pt]
Phase 2 --- Weibo Publishing: \\
Step 6: Open the Weibo compose page in the browser via web\_scan. \\
Step 7: Inject JavaScript to locate the post editor textarea and fill in the extracted message. \\
Step 8: Trigger the submit button via web\_execute\_js. \\
Step 9: Verify that the post appears on the user's Weibo timeline.
\end{CaseTraceBox}

\begin{CaseAnswerBox}{Observed Outcome}
\small\ttfamily
In this session, GA bridged two heterogeneous platforms within one workflow: \\
(1) local encrypted WeChat database $\rightarrow$ SQLCipher decryption $\rightarrow$ contact resolution $\rightarrow$ message extraction; \\
(2) browser-based Weibo $\rightarrow$ JS injection $\rightarrow$ post creation $\rightarrow$ publication verification. \\
The tools used were code\_run (SQLCipher operations, key derivation), file\_read (database file access), and web\_scan + web\_execute\_js (Weibo interaction). \\
This case specifically illustrates the paper's claim that atomic tools can be composed across heterogeneous local and web environments, allowing one session to move from private local data handling to browser-side execution under explicit user intent.
\end{CaseAnswerBox}

\FloatBarrier

\subsubsection{Case B3. Autonomous Operation: Unsupervised Overnight Session}
\label{app:case_autonomous}

\begin{CasePromptBox}{Purpose and Task Setup}
\small\ttfamily
Purpose: to examine whether GA can operate productively with limited oversight under a preset autonomy policy, \\
we present an autonomous session triggered by the system detecting that the user has been away for over 30 minutes. \\
This task is useful because it demonstrates self-directed task selection, execution, and self-correction over an extended period. \\
The goal of the case is to show that GA's self-evolution capability can extend beyond benchmark settings into bounded autonomous operation. \\[6pt]
Trigger condition: \\
{[AUTO]} User has been away for more than 30 minutes. \\
Operational boundary: the session was restricted to inspection, reporting, tool creation, and reversible local maintenance actions; external publication, payment, and destructive system modification were excluded by policy. \\
No explicit task instruction was given after the trigger; GA selected and executed tasks based on its built-in patrol and self-improvement routines within the preset policy.
\end{CasePromptBox}

\begin{CaseTraceBox}{Artifact 1: Autonomous Task Inventory (Rounds 125--140+)}
\small\ttfamily
Category 1 --- System Security Audit: \\
- Scanned all listening ports on the host machine. \\
- Discovered a stock-trading application occupying 939 UDP ports and 509\,MB of memory at 3:00\,AM. \\
- Audited startup items: 11 registry entries + 267 scheduled tasks + 86 auto-start services. \\[4pt]
Category 2 --- Tool Creation: \\
- Wrote port\_monitor.py: real-time port monitoring utility. \\
- Wrote process\_watchdog.py: process resource watchdog with alerting. \\
- Wrote startup\_auditor.py: startup-item auditing and reporting tool. \\[4pt]
Category 3 --- Web Patrol: \\
- Visited tech community sites (Guohe Boke, V2EX) and GitHub Trending. \\
- Generated a summarized patrol report covering trending topics and notable repositories. \\[4pt]
Category 4 --- Environment Hygiene: \\
- Audited the Python environment: found 294 installed packages occupying 1.9\,GB. \\
- Identified 2 CVE vulnerabilities in the filelock package; wrote pip\_audit.py. \\
- Discovered and fixed an inconsistency in its own history filenames (.md vs .txt extension mismatch).
\end{CaseTraceBox}

\begin{CaseAnswerBox}{Observed Outcome}
\small\ttfamily
Over approximately 15+ autonomous rounds, GA performed system security auditing, created three reusable utility scripts, conducted web patrol with report generation, audited the Python environment for vulnerabilities, and self-corrected its own file-management errors. \\
After the trigger fired, the session proceeded without additional user input and remained within the preset autonomy policy described above. \\
Tools used: code\_run (system commands, script creation and testing), file\_read / file\_write (script files, reports), web\_scan (web patrol), and update\_working\_checkpoint (tracking progress across rounds). \\
This case specifically illustrates the paper's claim that self-evolution, working checkpoints, and patrol routines can support bounded autonomous operation over extended periods, rather than only repeated benchmark-style tasks.
\end{CaseAnswerBox}

\FloatBarrier

\subsubsection{Case B4. Remote Infrastructure: SSH-Based File Server Deployment}
\label{app:case_remote_infra}

\begin{CasePromptBox}{Purpose and Task Setup}
\small\ttfamily
Purpose: to examine whether GA can manage end-to-end remote infrastructure tasks, \\
we provide a real-world deployment scenario requiring SSH access, file transfer, and iterative service configuration. \\
This task is useful because it involves dependency management on an unfamiliar remote machine, multi-step troubleshooting, and responding to changing user requirements mid-task. \\
The goal of the case is to show that an atomic-tool workflow can support a complete DevOps-style execution chain in this setting, without relying on infrastructure-as-code tooling. \\[6pt]
User instruction (verbatim, translated): \\
``SSH into user@{\it [IP address]}. Set up a file server on it and upload the zip file.'' \\
Follow-up instruction: ``Make it publicly accessible.''
\end{CasePromptBox}

\begin{CaseTraceBox}{Artifact 1: Execution Trace}
\small\ttfamily
Phase 1 --- Environment Setup: \\
Step 1: Install the paramiko library for SSH connectivity (pip install paramiko). \\
Step 2: Establish an SSH connection to the remote Linux server. \\
Step 3: Upload the 19\,MB zip file via SFTP. \\[4pt]
Phase 2 --- Service Deployment: \\
Step 4: Deploy a Python HTTP file server on the remote machine (python3 -m http.server). \\
Step 5: Encounter and fix a Chinese filename encoding issue (UTF-8 locale configuration). \\[4pt]
Phase 3 --- Requirement Change: \\
Step 6: User requests public access $\rightarrow$ reconfigure the server to remove authentication. \\
Step 7: Add directory isolation so that only the target folder is exposed. \\[4pt]
Phase 4 --- Verification: \\
Step 8: Test the download from the local machine. \\
Step 9: Confirm file integrity: downloaded size = 19.01\,MB, matching the original.
\end{CaseTraceBox}

\begin{CaseAnswerBox}{Observed Outcome}
\small\ttfamily
In this session, GA completed the full DevOps workflow: SSH connection, dependency installation on the remote machine, file upload (19\,MB), HTTP server deployment, encoding bugfix, mid-task requirement change (public access), directory isolation, and end-to-end verification. \\
Tools used: code\_run (paramiko SSH, SFTP, remote shell commands), file\_read / file\_write (local zip file, configuration), and update\_working\_checkpoint (phase tracking). \\
This case specifically illustrates the paper's claim that atomic tools plus lightweight state tracking can support a complete ``connect $\rightarrow$ configure $\rightarrow$ deploy $\rightarrow$ troubleshoot $\rightarrow$ verify'' workflow over SSH in a realistic remote environment.
\end{CaseAnswerBox}

\FloatBarrier

\subsubsection{Case B5. Long-Horizon Academic Workflow: NSFC Grant Proposal Assistance}
\label{app:case_academic}

\begin{CasePromptBox}{Purpose and Task Setup}
\small\ttfamily
Purpose: to examine whether GA can sustain coherent support over a long-horizon, multi-session academic task, \\
we present a real-world NSFC (National Natural Science Foundation of China) grant proposal assistance case that spanned multiple sessions over several days. \\
This task is useful because it involves reading and comprehending a full-length proposal, identifying structural weaknesses, producing academic figures, performing large-scale citation verification, and iteratively correcting errors --- capabilities that must be coordinated across sessions with persistent memory. \\
The goal of the case is to show that GA's memory system and self-evolution capabilities can support sustained, multi-phase academic assistance beyond a single session. \\[6pt]
Presentation note: the proposal materials were user-provided, and document-specific details are anonymized here when they are not essential to the technical point. \\
Task phases (emerged organically from user interaction): \\
1. Read and analyze the complete NSFC proposal document. \\
2. Identify that the proposal contains no figures; design a figure plan. \\
3. Generate academic figures (overview diagram + per-section illustrations). \\
4. Verify all bibliographic entries against external sources. \\
5. Correct discovered citation errors and regenerate the PDF.
\end{CasePromptBox}

\begin{CaseTraceBox}{Artifact 1: Figure Design and Generation}
\small\ttfamily
Observation: the entire proposal contained zero figures, which is a significant weakness for an NSFC application. \\
Action plan: \\
- Design a research overview figure showing the relationship among the three proposed research themes. \\
- Design per-section illustrations for each research content block. \\[4pt]
Execution: \\
- Used Python (matplotlib, networkx) to generate vector-format overview diagrams. \\
- Composed Gemini API prompts to generate domain-appropriate academic illustrations. \\
- Integrated figures into the LaTeX source and verified rendering.
\end{CaseTraceBox}

\begin{CaseTraceBox}{Artifact 2: Citation Verification and Batch Correction}
\small\ttfamily
Approach: parsed the .bib file and extracted all bibliographic entries. \\
For each entry: \\
- Queried arXiv API to verify title, authors, and year. \\
- Cross-referenced with OpenAlex API for published-venue entries. \\
- Compared extracted metadata against the .bib fields. \\[4pt]
Findings: \\
- Discovered a significant number of arXiv preprint entries with incorrect metadata (wrong year, mismatched titles, incomplete author lists). \\
- Root cause: likely auto-generated .bib entries from LLM-assisted writing without manual verification. \\[4pt]
Correction: \\
- Batch-updated the .bib file with verified metadata. \\
- Logged all changes for the user's review (change log with before/after comparisons). \\
- Regenerated the PDF with corrected references.
\end{CaseTraceBox}

\begin{CaseAnswerBox}{Observed Outcome}
\small\ttfamily
Across multiple sessions spanning several days, GA provided end-to-end assistance for: \\
(1) full-document comprehension and structural analysis; \\
(2) figure design and generation (Python vector graphics + API-generated illustrations); \\
(3) systematic citation verification against arXiv and OpenAlex, discovering and correcting a batch of erroneous entries; \\
(4) LaTeX integration and PDF regeneration. \\[4pt]
Tools used: file\_read (proposal document, .bib file), code\_run (Python plotting, API queries, .bib parsing), file\_write (corrected .bib, figures), web\_scan (arXiv / OpenAlex verification), and update\_working\_checkpoint (cross-session state). \\[4pt]
This case specifically illustrates the paper's claim that layered persistent memory, together with reusable procedures accumulated through self-evolution, can maintain continuity across multi-session academic workflows involving reading, generation, verification, and revision.
\end{CaseAnswerBox}

\FloatBarrier